%% file: Main.tex
\documentclass[lettersize,journal]{IEEEtran}
\usepackage{amsmath,amsfonts}
\usepackage{algorithmic}
\usepackage{array}
\usepackage[caption=false,font=normalsize,labelfont=sf,textfont=sf]{subfig}
\usepackage{textcomp}
\usepackage{stfloats}
\usepackage{url}
\usepackage{verbatim}
\usepackage{graphicx}
\usepackage{hyperref}
\usepackage{multirow}

\def\BibTeX{{\rm B\kern-.05em{\sc i\kern-.025em b}\kern-.08em
    T\kern-.1667em\lower.7ex\hbox{E}\kern-.125emX}}
\usepackage{balance}
\urldef\Ottourl\url{https://shop.ottobock.us/Prosthetics/Upper-Limb-Prosthetics/Myoelectric-Elbows/DynamicArm/p/12K100N%7E550-1.com}

\begin{document}
\title{Design, Characterization, and Validation \\of a Variable Stiffness Prosthetic Elbow}
\author{Giuseppe Milazzo, Simon Lemerle, Giorgio Grioli, Antonio Bicchi, and Manuel G. Catalano
\thanks{Manuscript first submitted on March, 2024; Accepted on October, 2024. This research has received funding from the European Union’s ERC Synergy Grant Agreement (810346) Natural BionicS, H2020 Programme under grant agreement RePAIR (964854), and by the Italian Ministry of Education and Research (MIUR) in the framework of the FoReLab project (Departments of Excellence). The content of this publication is the sole responsibility of the authors. The European Commission or its services cannot be held responsible for any use that may be made of the information it contains. 
Corresponding author: Giuseppe Milazzo. e-mail: giuseppe.milazzo@iit.it.
G. Milazzo, G. Grioli, A. Bicchi, and M.G. Catalano are with Soft Robotics for Human Cooperation and Rehabilitation, Istituto Italiano di Tecnologia, Genova 16163, Italy. 
G. Grioli and A. Bicchi are also with the Research Center “Enrico Piaggio” \& Department of Information Engineering, University of Pisa, Pisa 56126, Italy. 
S. Lermerle was formerly with all the above-mentioned affiliations.
This paper has supplementary downloadable material available at \url{http://ieeexplore.ieee.org}}}

\markboth{Journal of \LaTeX\ Class Files,~Vol.~18, No.~9, September~2020}%
{How to Use the IEEEtran \LaTeX \ Templates}

\maketitle
\begin{abstract}
Intuitively, prostheses with user-controllable stiffness could mimic the intrinsic behavior of the human musculoskeletal system, promoting safe and natural interactions and task adaptability in real-world scenarios. 
However, prosthetic design often disregards compliance because of the additional complexity, weight, and needed control channels. 
This paper focuses on designing a Variable Stiffness Actuator (VSA) with weight, size, and performance compatible with prosthetic applications, addressing its implementation for the elbow joint.
While a direct biomimetic approach suggests adopting an Agonist-Antagonist (AA) layout to replicate the biceps and triceps brachii with elastic actuation, this solution is not optimal to accommodate the varied morphologies of residual limbs.
Instead, we employed the AA layout to craft an elbow prosthesis fully contained in the user's forearm, catering to individuals with distal transhumeral amputations.
Additionally, we introduce a variant of this design where the two motors are split in the upper arm and forearm to distribute mass and volume more evenly along the bionic limb, enhancing comfort for patients with more proximal amputation levels.
We characterize and validate our approach, demonstrating that both architectures meet the target requirements for an elbow prosthesis. The system attains the desired 120° range of motion, achieves the target stiffness range of [2, 60] Nm/rad, and can actively lift up to 3 kg. Our novel design reduces weight by up to 50\% compared to existing VSAs for elbow prostheses while achieving performance comparable to the state of the art. Case studies suggest that passive and variable compliance could enable robust and safe interactions and task adaptability in the real world.
\end{abstract}
\input{Sections/Introduction}
\input{Sections/DesignConcept}
\input{Sections/WorkingPrinciples}
\input{Sections/Mechatronic}
\input{Sections/Control}
\input{Sections/Experiments}
\input{Sections/Discussions}
\input{Sections/Conclusions}
\section*{Acknowledgements}
The authors would like to thank Manuel Barbarossa, Mattia Poggiani, Marina Gnocco, and Emanuele Sessa for their fundamental technical support.
\bibliographystyle{ieeetr}
\bibliography{References}

\begin{IEEEbiography}[{\includegraphics[width=1in,height=1.25in,clip,keepaspectratio]{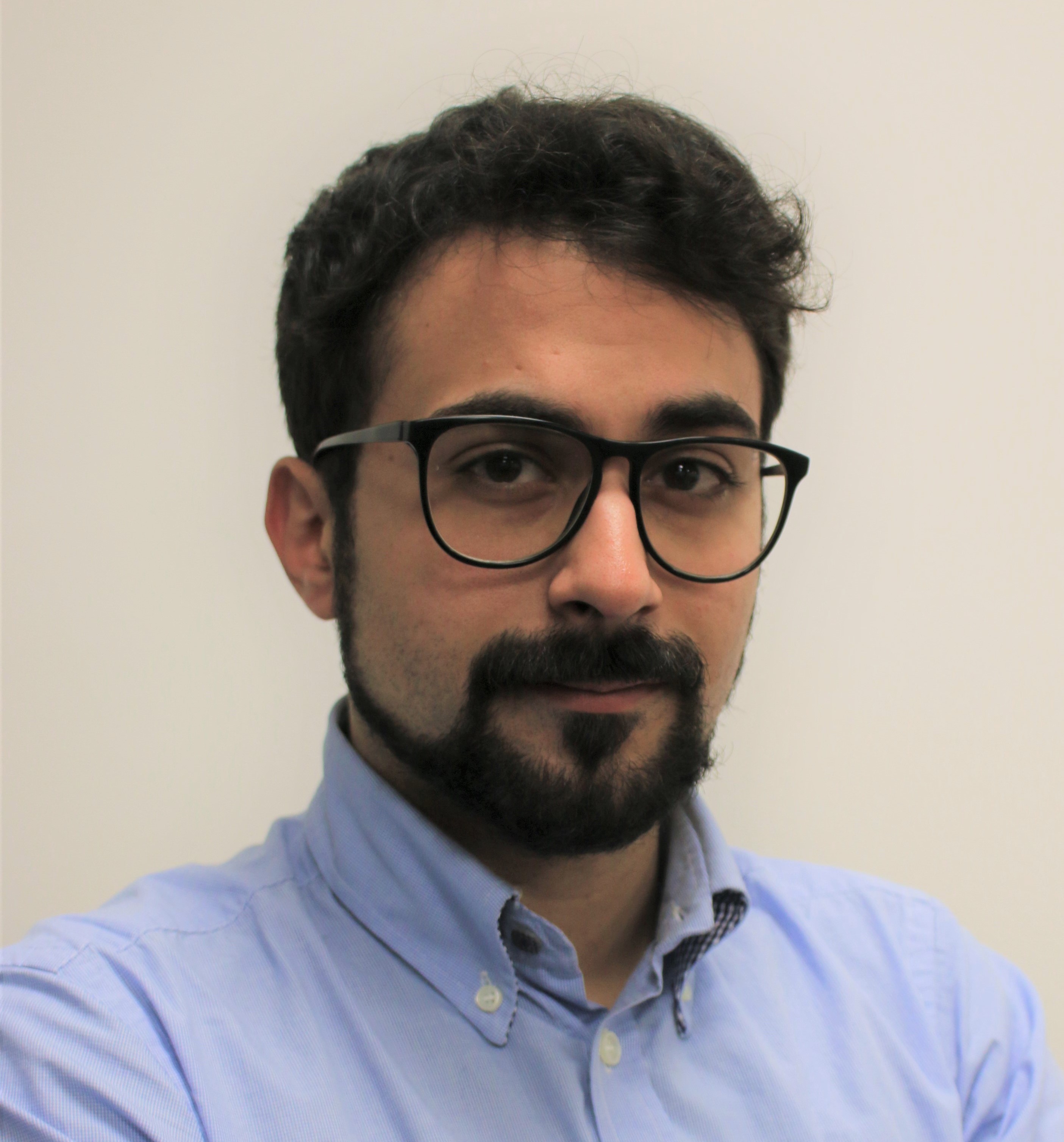}}]{Giuseppe Milazzo} received the B.S. degree \textit{cum laude} in biomedical engineering and the M.S. degree \textit{cum laude} in robotics and automation engineering from University of Pisa, Pisa, Italy.

He is currently a Ph.D. candidate for the national doctorate in robotics and intelligent machines, conducting his research activities at the Soft Robotics for Human Cooperation and Rehabilitation laboratory, Italian Institute of Technology, Genova, Italy.
His main research interests include design, modeling, and control of Variable Stiffness Actuators and upper limb prostheses. 
\end{IEEEbiography}

\begin{IEEEbiography}
[{\includegraphics[width=1in,height=1.25in,clip,keepaspectratio]{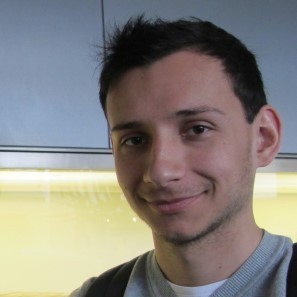}}]{Simon Lemerle} received in 2017 the Engineer's degree from the Ecole Centrale de Lyon, Lyon, France, and the M.E. degree in integrated design engineering from Keio University, Yokohama, Japan. He received the Ph.D. degree in Robotics from the University of Pisa, Pisa, Italy in 2021, working on soft robotics for upper limb prosthesis in the Department of Information Engineering, University of Pisa, Pisa, Italy, and the Italian Institute of Technology, Genoa, Italy. His research interests include haptics, upper-limb robotics, and rehabilitation.
\end{IEEEbiography}

\begin{IEEEbiography}
[{\includegraphics[width=1in,height=1.25in,clip,keepaspectratio]{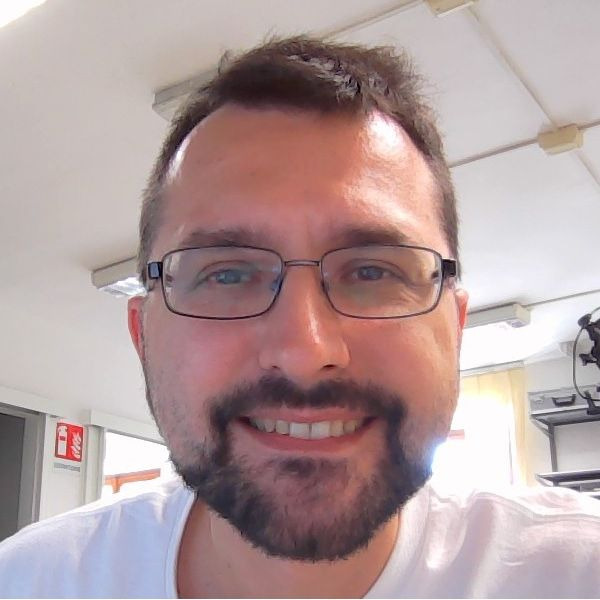}}]{Giorgio Grioli} received the Ph.D. degree in robotics and automation engineering from the University di Pisa, Pisa, Italy, in 2011.

He is a Researcher with the Istituto Italiano di Tecnologia, Genova, Italy, and investigates the design,
modeling, and control of soft robotics systems applied
to augmentation of, rehabilitation of and interaction
with the human.
He authored more than 120 journal articles, conference articles, and book chapters (4,500 citations, h-index 29). He is reported as inventor in six between patent and patent applications. He contributed to the development of several robotic systems and to the founding of a spin-off company. His research interest includes collaborative industrial robotics to prosthetics.
\end{IEEEbiography}

\begin{IEEEbiography}
[{\includegraphics[width=1in,height=1.25in,clip,keepaspectratio]{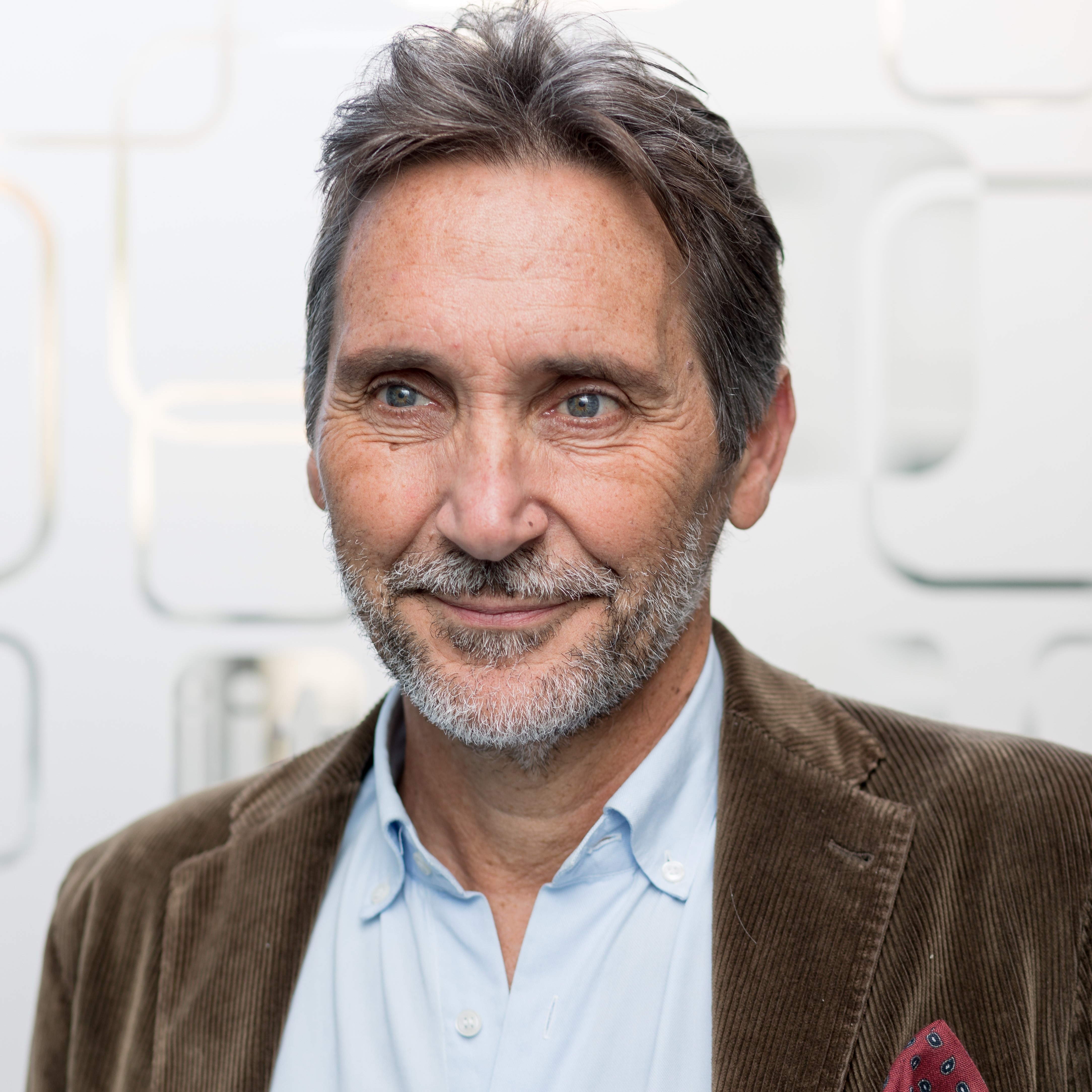}}]{Antonio Bicchi} received the Ph.D. degree in mechanical engineering from the University of Bologna, Bologna, Italy. He is a Scientist interested in robotics and intelligent machines. 

He holds a chair in Robotics with the University of Pisa, Pisa, Italy, and leads the Soft Robotics Laboratory, Italian Institute of Technology, Genova, Italy. He is also an Adjunct Professor with Arizona State University, Tempe, USA. His work has been recognized with many international awards and has earned him four prestigious grants from the European Research Council (ERC). He launched initiatives such as the WorldHaptics conference series, the IEEE Robotics and Automation Letters, and the Italian Institute of Robotics and Intelligent Machines.
\end{IEEEbiography}

\begin{IEEEbiography}
[{\includegraphics[width=1in,height=1.25in,clip,keepaspectratio]{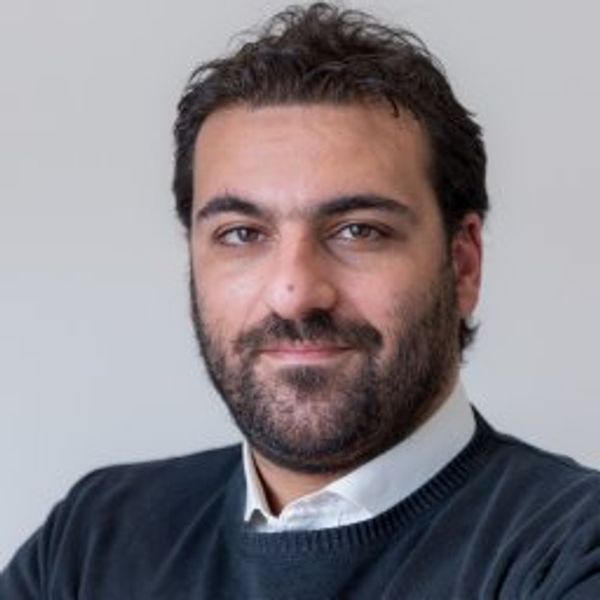}}]{Manuel Giuseppe Catalano} received the B.S. and M.S. degrees in mechanical engineering and the Ph.D. degree in robotics from the University of Pisa, Pisa, Italy, in 2006, 2008, and 2013, respectively.

He is currently a Researcher with the Italian Institute of Technology, Genoa, Italy, and a collaborator of the Research Center “E. Piaggio,” University of Pisa. His research interests include the design of soft robotic systems, human-robot interaction, and prosthetics.

Dr. Catalano was the recipient of the Georges Giralt Ph.D. Award in 2014, the prestigious annual European award given for the best Ph.D. thesis by euRobotics AISBL.
\end{IEEEbiography}

\end{document}

%% file: Sections/Introduction.tex
\section{Introduction}\label{sect:Introduction}
\IEEEPARstart{M}{issing} a limb profoundly impacts individuals' physical, emotional, and social well-being.
Prosthetic devices seek to address these issues by replicating the natural capabilities and appearance of human limbs.
However, current prostheses are still far from achieving the functionality of their natural counterparts and rather focus on restoring users' auto-sufficiency with rudimental motor capabilities.
Most upper limb prosthetic research concentrates on hands as they serve as the primary interface for interacting with the environment. Nevertheless, approximately 21\% of upper limb loss occurs above the elbow joint \cite{cordella2016literature}, underscoring the need to address motor functions at the elbow and wrist levels.

The development of mechanical systems replicating human characteristics typically focuses on shape, power, and kinematics, overlooking their behavior during physical interactions. 
Nevertheless, aligning the impedance of natural and artificial limbs during the design process would achieve robust, safe, and natural interactions within unknown and dynamic environments \cite{perreault2014considering}.
Humans modulate limb impedance through muscular cocontraction, exploiting a compliant behavior to minimize interaction forces in constrained tasks or explore unknown environments while increasing stiffness to accomplish assignments that demand precision or resisting external perturbations \cite{osu2004optimal,borzelli2018muscle,blank2014task}. Therefore, upper limb prostheses should incorporate user-controllable impedance \cite{hogan1983prostheses}.

\begin{figure}[!t]
    \centering
    \subfloat[]{\includegraphics[width = \linewidth]{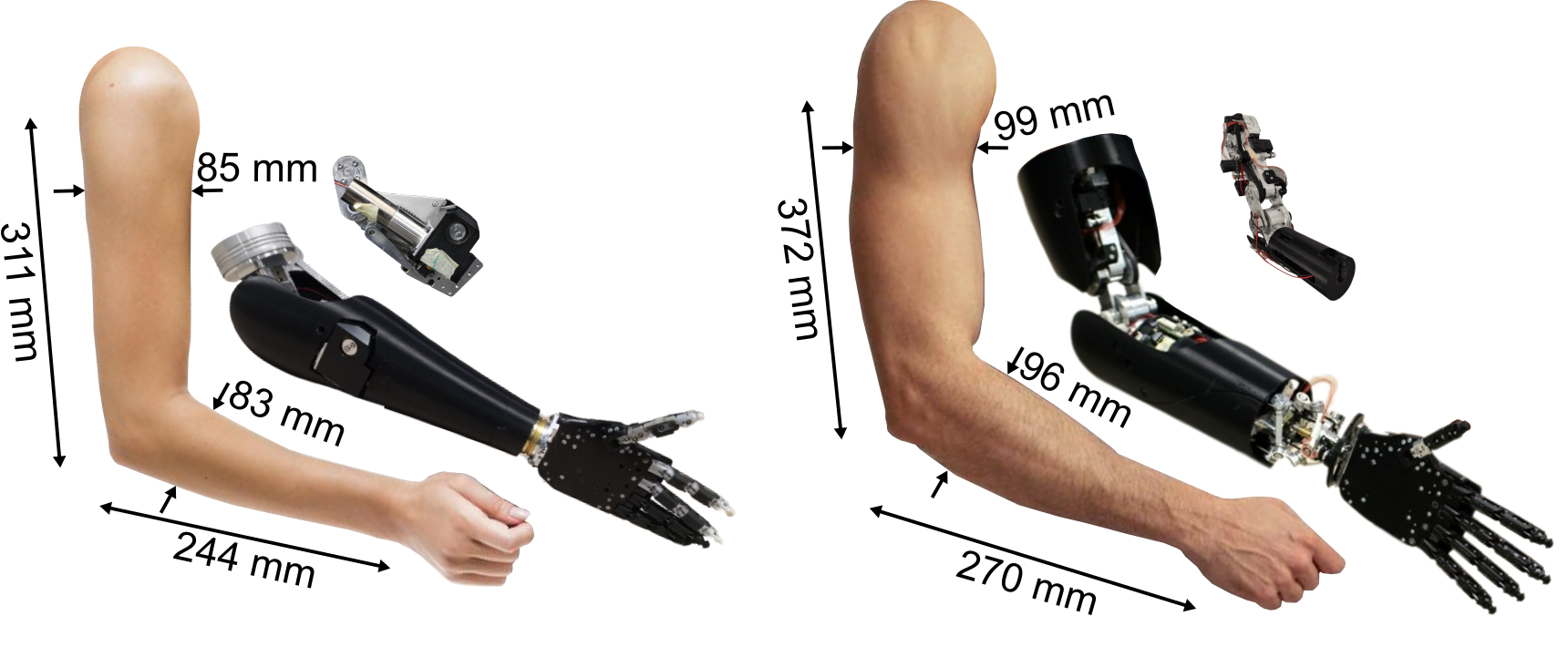}\label{fig:Fig1a}}\\
    \subfloat[]{\includegraphics[width = \linewidth]{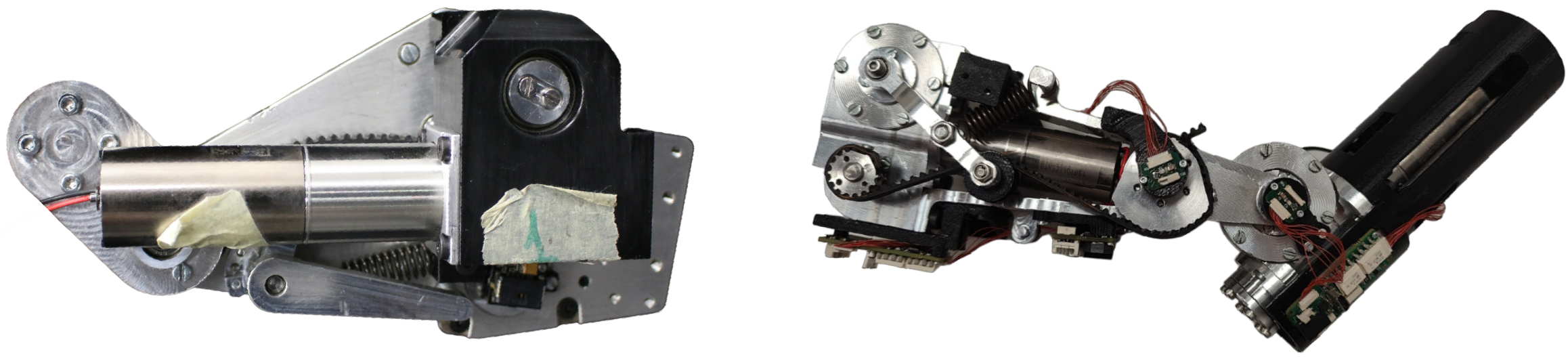}}
    \caption{The presented variable stiffness elbow and distinct applications in transhumeral prostheses. (a) The bionic limb on the left adopts the VS-Elbow AA to locate all components below the elbow joint, thus being suitable for distal transhumeral amputations. The prosthesis on the right employs the VS-Elbow D2 to achieve a sparser mass distribution, enhancing user comfort in more proximal transhumeral amputations. Panel (b) highlights the two implementations of the variable stiffness elbow: the VS-Elbow AA on the left and the VS-Elbow D2 on the right.}
    \label{fig:Fig1}
\end{figure}
A natural approach to controlling variable impedance prostheses correlates their stiffness with antagonistic muscle cocontraction and motion with the difference in muscular activity measured by the user's electromyography (EMG) signals \cite{hogan1983prostheses, capsi2020exploring}. A more sophisticated method may be to adopt EMG-driven muscle-tendon models \cite{ferrante2024toward}.
Sensinger et al. investigated the applicability of EMG-driven active impedance control of an elbow joint, concluding that users do not deliberately adjust the impedance level during reaching tasks if around a suitable baseline \cite{sensinger2008user}. 
More recently, Blank et al. proved that preferred impedance levels are task-dependent \cite{blank2014task}, implying potential benefits for user-modulated impedance in prosthetic limbs and highlighting the need for more effective EMG control strategies.

Variable impedance can be implemented in robots actively through software control \cite{hogan1985impedance} or passively by embedding reconfigurable elastic elements in the mechanical design \cite{grioli2015variable}.
However, software impedance regulation requires accurate sensing of the output state and fast actuation and computation, which are unlikely to be achieved in prostheses due to weight and size constraints \cite{english1999mechanics}.
Conversely, passive impedance regulation is intrinsically robust, as the embedded elastic elements inherently absorb shocks and store energy, potentially improving performance in explosive or cyclic tasks and reducing power consumption \cite{albu2008soft, vanderborght2006exploiting, haddadin2011optimal, haddadin2007safety}, \cite{govindan2020design}. Nevertheless, this approach leads to more complex mechanisms, requiring an additional actuator to reconfigure the compliant elements and achieve stiffness modulation. Consequently, the resulting increase in size and weight poses challenges in meeting the morphological design requirements of prostheses.

Commercial elbow prostheses\footnote{\label{note1}\href{https://www.utaharm.com/product-utah-arm-options/}{Fillauer Utah Arm U3}}${}^,$\footnote{\label{note2}\href{https://shop.ottobock.us/Prosthetics/Upper-Limb-Prosthetics/Myoelectric-Elbows/DynamicArm/p/12K100N~550-1}{Ottobock Dynamic Arm}} feature rigid actuation that can be locked to hold external weights, proportionally controlled for active motion, or unlocked to swing freely, minimizing effort in lowering the arm and improving walking performance, by reducing metabolic cost and enhancing gait stability \cite{collins2009dynamic}, \cite{bruijn2010effects}.
Even research prototypes conventionally adopt rigid actuation \cite{abayasiri2017mobio,lenzi2016ric, bennett2016design}, and rarely implement active impedance control \cite{sensinger2008user}. However, the literature also contains a few examples of employing passive impedance regulation in prosthetic devices, such as ankle-foot \cite{shepherd2017vspa, glanzer2018design, rogers2023design}, hand \cite{hocaoglu2022design}, wrist \cite{milazzo2024modeling}, and elbow \cite{lemerle2019variable, baggetta2022design} prostheses.
In \cite{lemerle2019variable}, the authors present a prosthetic elbow featuring a VSA that varies joint stiffness by modulating the preload of linear springs. However, the first prototype design was bulky and has been optimized in the presented work.
In \cite{baggetta2022design}, Baggetta et al. propose a similar architecture achieving stiffness modulation via 3D-printed torsion springs, whose geometry has been optimized to fit a desired non-linear behavior.
However, the tested range of joint stiffness and torque fell substantially below the functional requirements of the human elbow.

This manuscript focuses on the design, characterization, and validation of a prosthetic elbow achieving Variable Stiffness (VS) through non-linear and redundant elastic actuation, expanding the work introduced in \cite{lemerle2019variable}.
We present a generic approach to the design of VS elbows and two design instances aiming to deliver a prosthesis tailored to accommodate diverse user needs and residual limb morphologies. In addition to \cite{lemerle2019variable}, we present the VS-Elbow AA, a novel hardware designed with an Agonist-Antagonist layout to accommodate lengthy residual limbs. Here, we show its implementation in a complete transhumeral prosthesis that can be fully contained in the user’s forearm. Moreover, we present the VS-Elbow D2, which is a more optimized and compact version of the prototype introduced in \cite{lemerle2019variable}, implementing a distributed design to achieve uniform weight allocation. 
Section~\ref{sect:Req} details the system target requirements, and Section~\ref{sect:design_conc} provides an overview of the employed architectures. Section~\ref{sect:work_princ} describes the working principles of the system, while the mechatronic design is detailed in Section~\ref{sect:Mechatronic Design}. Section~\ref{sect:Control} reports the implemented control strategy. Section~\ref{sect:Experimental Protocol} details the experimental protocol adopted to characterize the platform and presents some case studies proving the effectiveness of VS in unstructured environments. Results are presented in Section~\ref{sect:results} and discussed in Section~\ref{sec:discussions}. Finally, conclusions are drawn in Section~\ref{sec:conclusions}.

%% file: Sections/DesignConcept.tex
\section{Design Requirements} \label{sect:Req}
\begin{table}[!t]
\centering
\caption{Morphological and functional requirements of a prosthetic elbow. Morphological requirements are set to match the anthropometrical data of the 50th percentile American male (M) and female (F) extracted from \cite{milazzo2024modeling,nasa_dim}. Functional requirements are set to accomplish standard ADLs, taking \cite{gates2016range, buckley1996dynamics, murray2004study, Popescu2003Elbow} as baseline.}
\label{tab:requirements}
\renewcommand{\arraystretch}{1.5}
\begin{tabular}{c | c  c | c  c | c | c}
\hline
\multicolumn{5}{c|}{\textbf{Morphological Requirements}} & \multicolumn{2}{c}{\begin{tabular}{c}
     \textbf{Functional  Requirements}
\end{tabular}} \\
\hline
 & \multicolumn{2}{c|}{U-Segment} & \multicolumn{2}{c|}{F-Segment} & \multicolumn{2}{c}{Elbow Joint}\\
\cline{2-7}
 & M & F & M & F & RoM (°) & [0,120] \\
\cline{1-5}
L (mm) & 183 & 149 & 135 & 122 & $\omega$ (°/s) & $\leq$ 250 \\
D (mm) & 99 & 81 & 98 & 83 & $\tau$ (Nm) & [-2.8, 5.9] \\
m (g) & 1250 & 940 & 725 & 545 & $\sigma$ (Nm/rad) & [2, 60] \\
\hline
\multicolumn{7}{c}{\begin{tabular}{@{}c@{}}  \footnotesize L = segment length; D = segment diameter; m = segment mass; \\ $\omega$ = elbow speed; $\tau$ = elbow torque; $\sigma$ = elbow stiffness. \end{tabular}}\\
\end{tabular}
\end{table}
Prosthetic design must adhere to specific constraints related to the overall shape, dimensions, and weight of the system to suit the prosthetic application. Additionally, to effectively replace a human elbow, the device must meet minimum requirements in terms of ranges of motion, torque, and stiffness.
The presented system consists of two segments connected by the elbow joint: the U-segment (upper arm side) and the F-segment (forearm side). Size and weight specifications are derived from \cite{nasa_dim} and \cite{milazzo2024modeling}. The morphological requirements of the U and F segments are reported in Table~\ref{tab:requirements} and set at half of the size and weight of the human upper arm and forearm, respectively.
The required elbow range of motion (RoM) to accomplish activities of daily living (ADLs) is approximately 0° to 120° \cite{gates2016range, buckley1996dynamics}. The recorded maximum torque is 5.8 Nm for flexion (while lifting a 2 kg mass to head height) and -2.8 Nm for extension (during personal hygiene tasks) \cite{murray2004study}. Maximum elbow speed during ADLs is observed at around 250 °/s for flexion during drinking and 126 °/s during eating with a spoon \cite{buckley1996dynamics}.
The range of elbow stiffness exploited by humans during ADLs remains undetermined. Zawadzki et al. state that elbow stiffness during cyclic movements is directly proportional to the frequency and inversely proportional to the amplitude of the cycle, ranging from 15 to 130 Nm/rad \cite{zawadzki2010maximal}. 
Popescu et al. demonstrate that, in reaching tasks, the maximum stiffness can significantly vary among different subjects, with the best fit being around 60 Nm/rad, while the stiffness of the relaxed limb is approximately 2 to 10 Nm/rad \cite{Popescu2003Elbow}.

The system requirements summarized in Table~\ref{tab:requirements} serve as a baseline for crafting a VS elbow prosthesis.
Nevertheless, it is desirable to reduce size and weight to enhance comfort and appearance and minimize the load on the user's proximal joints during ADLs, while increasing torque and speed to improve performance. 
Reduced prosthesis weight has emerged as consumers' highest priority design concern, followed by lower cost, life-like appearance, and wearing comfort \cite{biddiss2007consumer, cordella2016literature}. Finally, the joint impedance should be designed in relation to the device mass to achieve human-like interactions and passive arm swinging during walking.

\section{System Concept}\label{sect:design_conc}
\begin{figure}[!t]
    \centering   
    \subfloat[]{\label{fig:2a}\includegraphics[width = 0.32\linewidth]{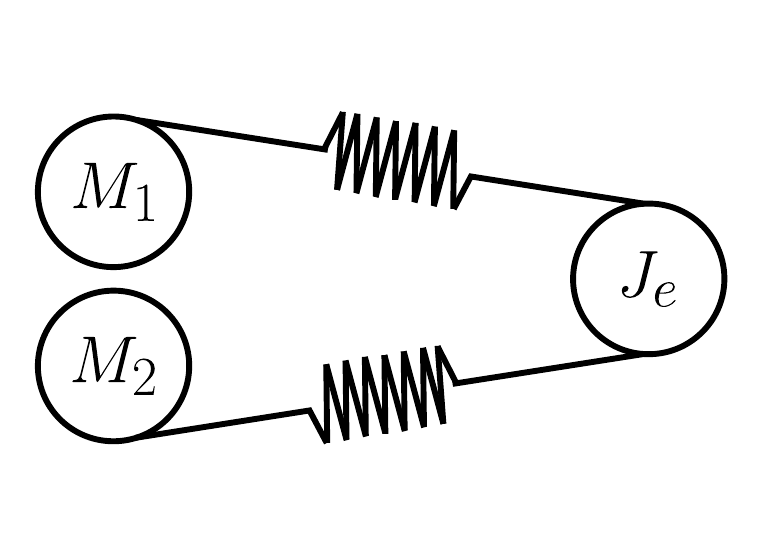}}
    \hfill
    \subfloat[]{\label{fig:2b}\includegraphics[width = 0.32\linewidth]{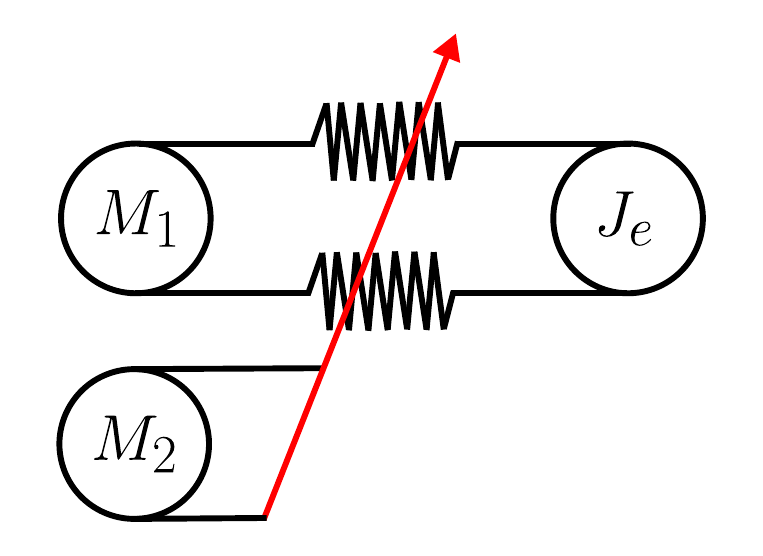}}
    \hfill
    \subfloat[]{\label{fig:2c}\includegraphics[width = 0.32\linewidth]{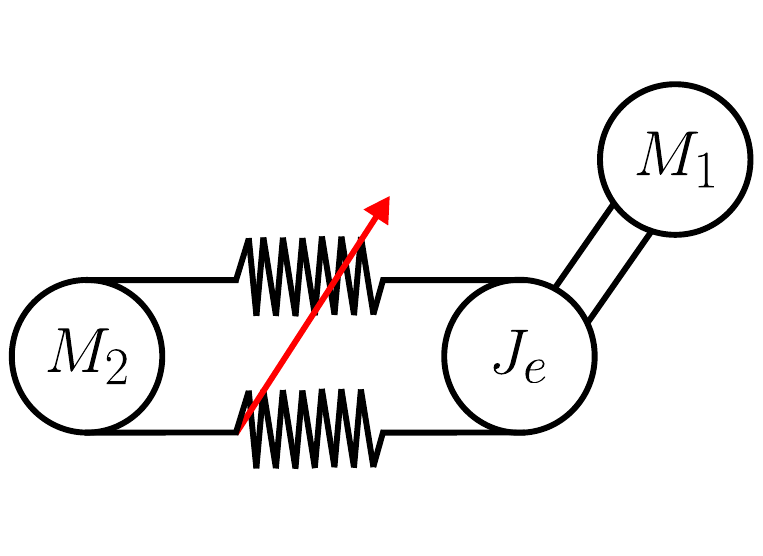}}\\
    \subfloat[]{\label{fig:2d}\includegraphics[width = 0.32\linewidth]{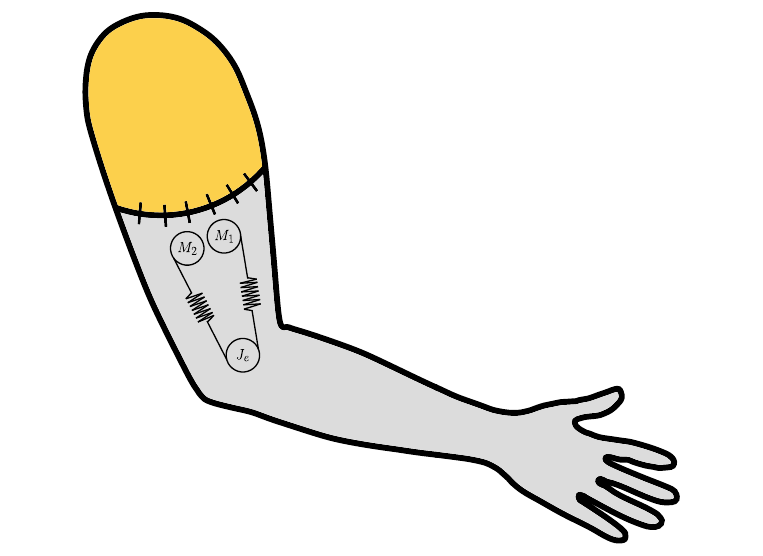}}
    \hfill
    \subfloat[]{\label{fig:2e}\includegraphics[width = 0.32\linewidth]{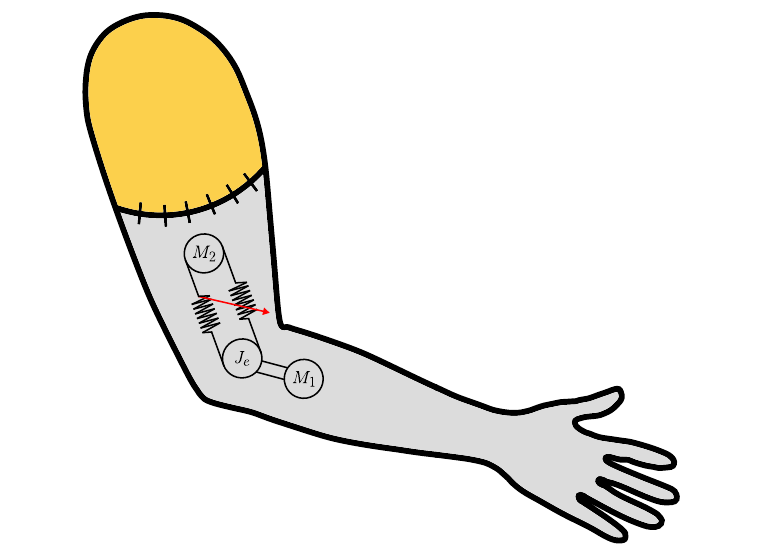}}
    \hfill
    \subfloat[]{\label{fig:2f}\includegraphics[width = 0.32\linewidth]{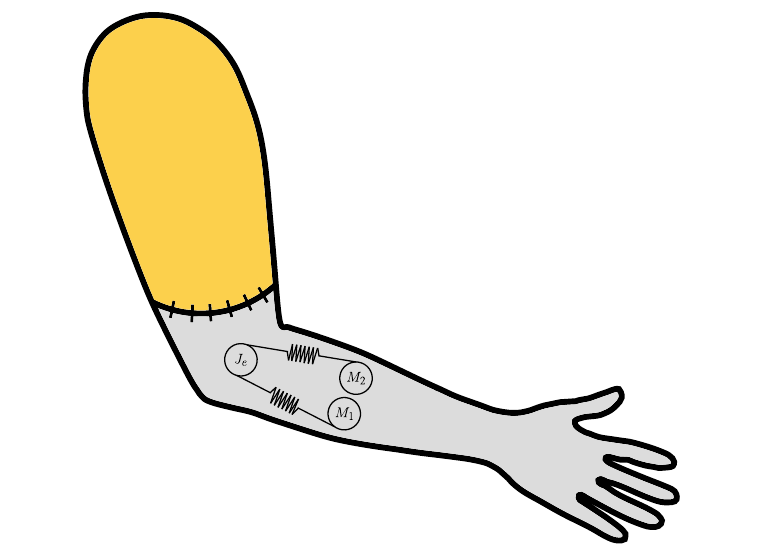}} 
    \caption{Schematic representation of diverse VSA architectures and their potential implementation in a VS elbow prosthesis. All the presented architectures feature a non-linear elastic transmission and two motors ($M_1, M_2$) to modulate the elbow joint ($J_e$) stiffness and position by exploiting redundant actuation.  Scheme (a) represents the Agonist-Antagonist architecture, whose functioning is inspired by the human musculoskeletal system. Scheme (b) displays the classical implementation of the independent setup, where each motor independently regulates either the position or the stiffness of the elbow joint. Scheme (c) illustrates a novel layout of the independent setup shown in (b), where the position and stiffness motors are placed on opposite elbow segments. 
    Panel (d) shows scheme (a) most intuitive implementation, where the elastic actuation is placed in the upper arm to replace the biceps and triceps brachii biomimetically.
    The implementation represented in panel (e) leverages scheme (c) to achieve homogeneous mass distribution, thus enhancing user comfort. Panel (f) displays the alternative fitting of the AA implementation that allocates all the components in the user's forearm, thus being suitable for distal transhumeral amputations.}
    \label{fig:Fig2}
\end{figure}
Controlling the position and stiffness of a VSA requires at least two motors. However, this goal can be achieved with different actuation layouts. The two main configurations are the agonist-antagonist and the independent setups \cite{kizilhan2015comparison}. The AA architecture, shown in Fig. \ref{fig:2a}, is inspired by the human muscles and consists of two opposed actuation systems, each comprising a motor unit and a non-linear elastic transmission to convey the motion to the output joint \cite{migliore2005biologically, moore2021design}. 
Similar to the human musculoskeletal system, the elastic actuators operate in an AA configuration, applying two opposing torques to the output joint to control its motion and stiffness.
Given that the elastic mechanisms simulate the biceps and triceps brachii, it would be natural to place both of them in the proximal part of the joint, as shown in Fig.~\ref{fig:2d}. 
In the independent configuration, illustrated in Fig. \ref{fig:2b}, one motor drives the output position, while the other adjusts the stiffness of the joint independently \cite{english1999mechanics}.
For simplicity, in this configuration as well, all components are typically placed on the same side of the joint. However, this approach challenges both user comfort and the fulfillment of morphological requirements.
To achieve an anthropomorphic shape and mass distribution, we developed an architecture based on the independent setup with a sparse allocation of the actuators along the system.
Its schematization, portrayed in Fig. \ref{fig:2c}, is equivalent to the one in Fig.~\ref{fig:2b}, considering $M_1$ in the inertia of the output link. Therefore, the implementation shown in Fig.~\ref{fig:2e} evenly distributes the device mass and volume across the forearm and upper limb segments, resulting in a reduction of the overall width and improved user comfort. 
However, the layouts portrayed in Fig. \ref{fig:2d} and Fig.~\ref{fig:2e} could be unsuitable for fitting lengthy residual limbs as the prosthetic elbow rotation axis may stand significantly below the contralateral elbow. 
In this situation, all components should fit in the user's forearm, as shown in Fig.~\ref{fig:2f} and Fig.~\ref{fig:Fig1a} for the AA layout.
Catering to diverse users' needs, this work implements the discussed architectures, presenting: the VS-Elbow AA, featuring the compact AA configuration shown in Fig.~\ref{fig:2f}, and the VS-Elbow D2, which adopts the distributed independent implementation illustrated in Fig.~\ref{fig:2e}.

%% file: Sections/WorkingPrinciples.tex
\section{Working Principles}\label{sect:work_princ}
\begin{figure*}[!t]
    \centering
    \subfloat[]{\includegraphics[width = 0.45\linewidth]{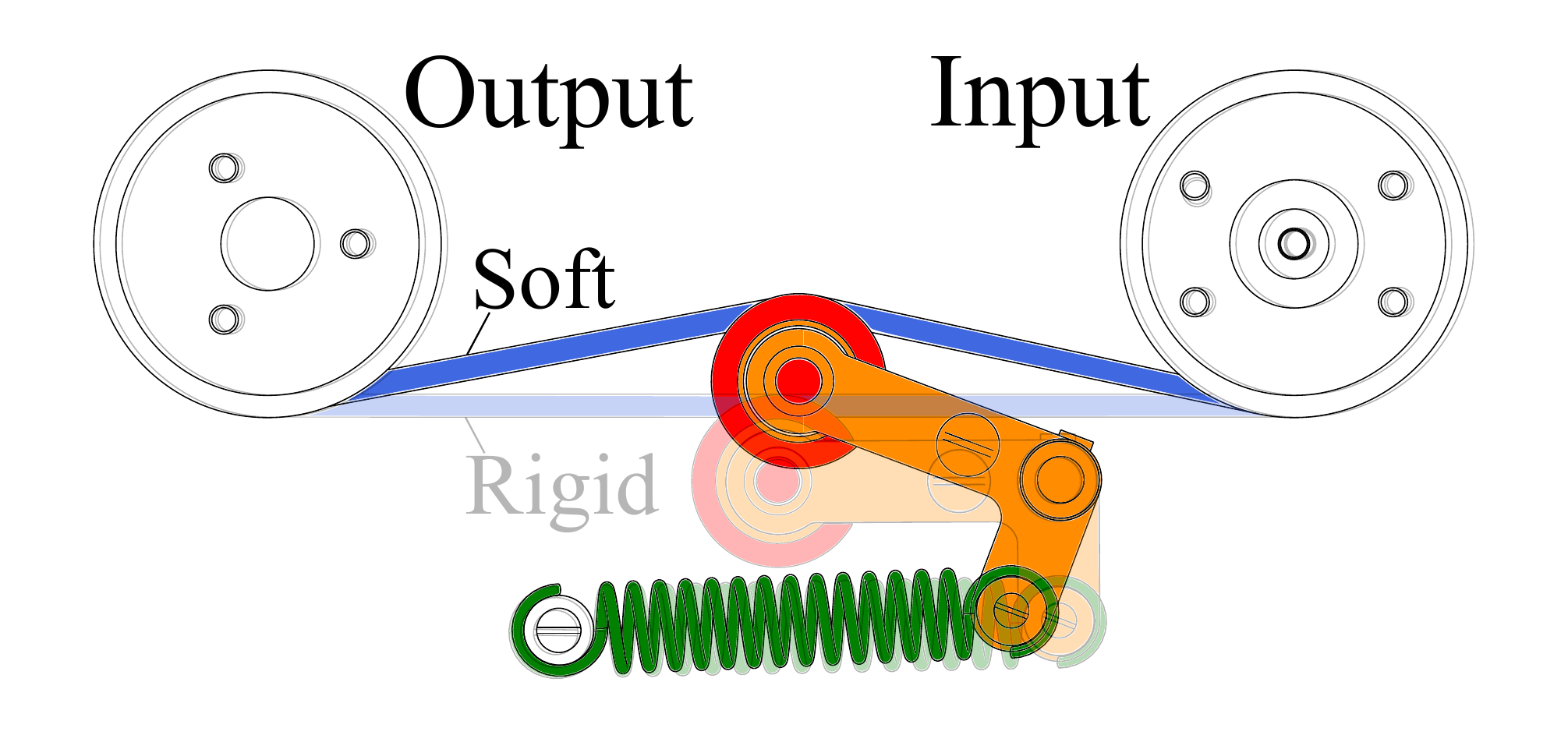}\label{fig:WorkingPrinciplesNew/3a}}
    \hfill    
    \subfloat[]{\includegraphics[width = 0.45\linewidth]{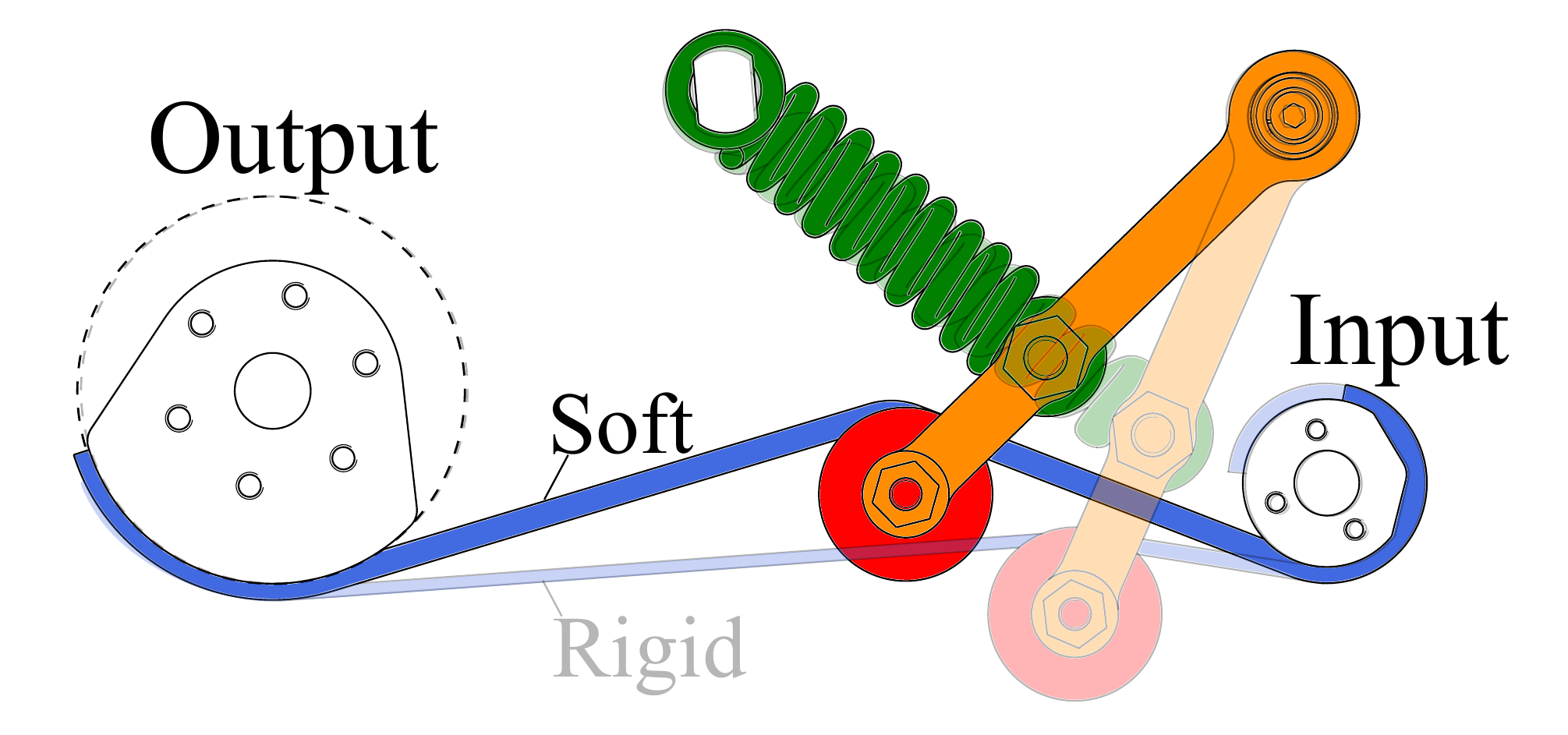}\label{fig:WorkingPrinciplesNew/3b}}\\
    \subfloat[]{\includegraphics[width = 0.3\linewidth]{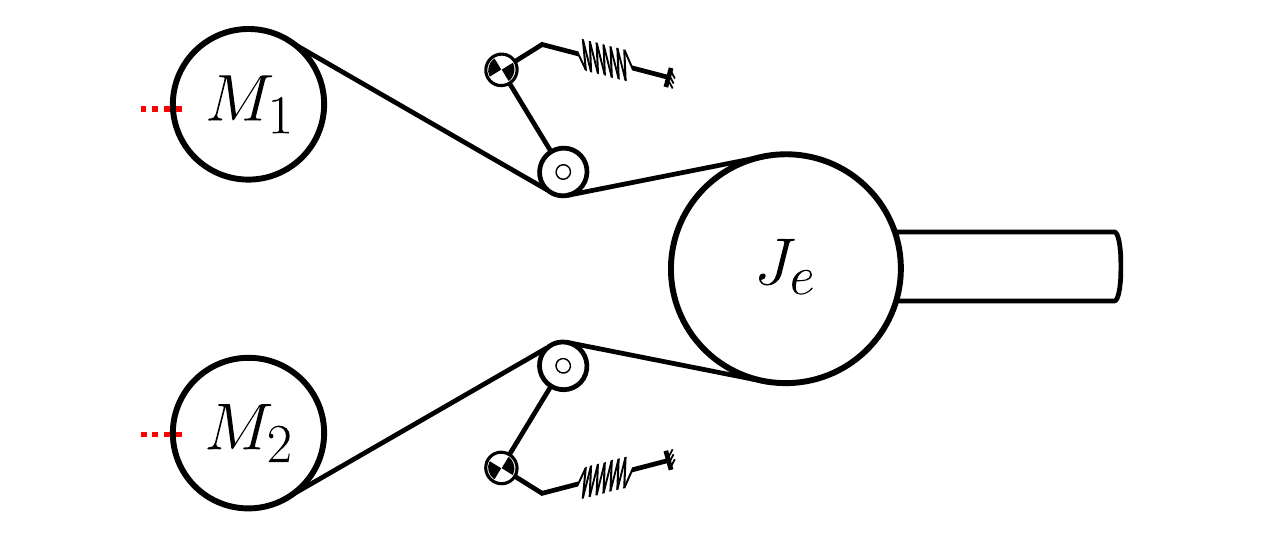}\label{fig:WorkingPrinciplesNew/3c}}
    \hfill    
    \subfloat[]{\includegraphics[width = 0.3\linewidth]{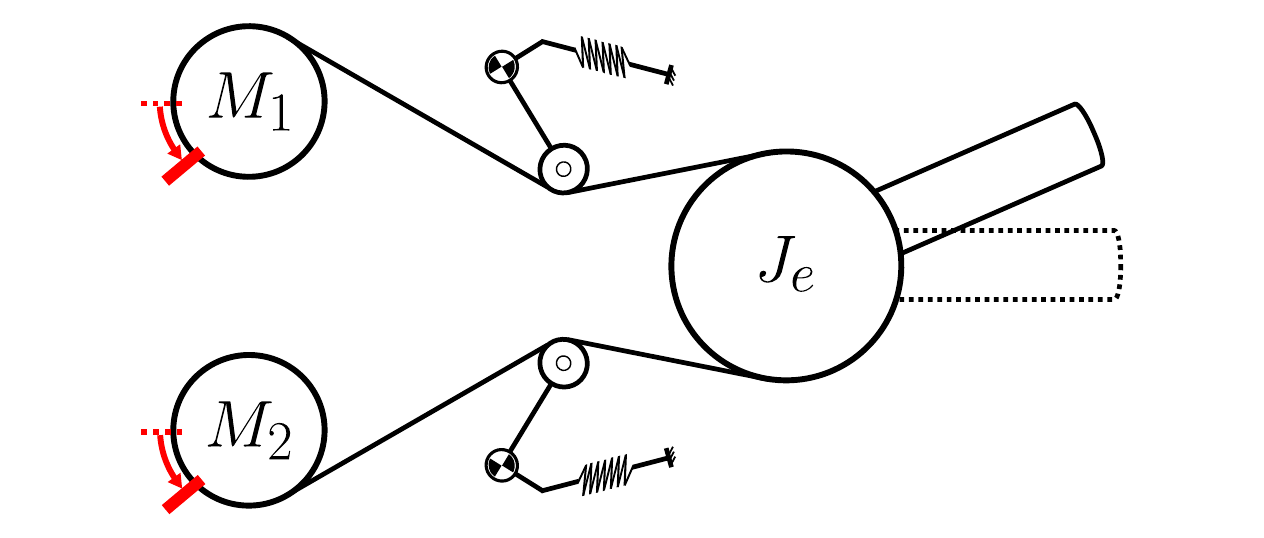}\label{fig:WorkingPrinciplesNew/3d}}
    \hfill    
    \subfloat[]{\includegraphics[width = 0.3\linewidth]{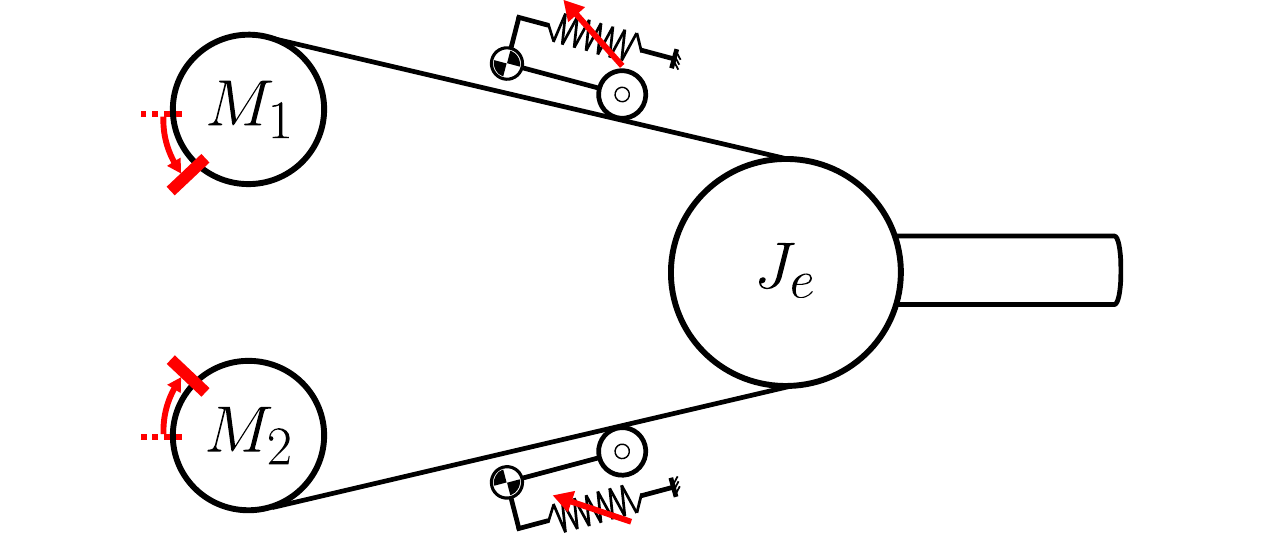}\label{fig:WorkingPrinciplesNew/3e}}\\
    \subfloat[]{\includegraphics[width = 0.3\linewidth]{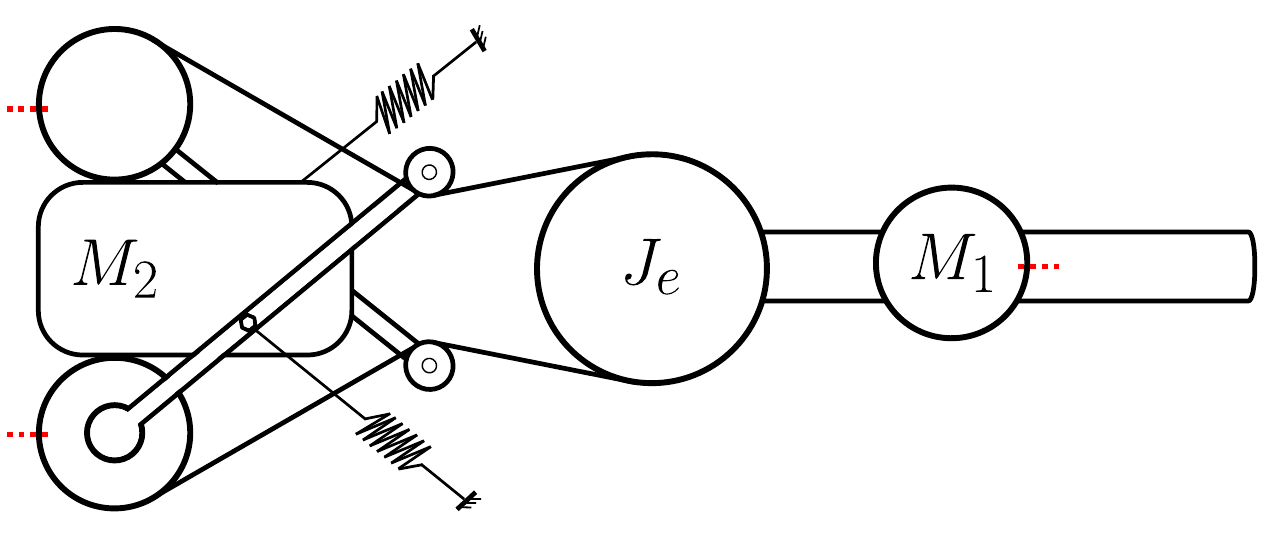}\label{fig:WorkingPrinciplesNew/3f}}
    \hfill    
    \subfloat[]{\includegraphics[width = 0.3\linewidth]{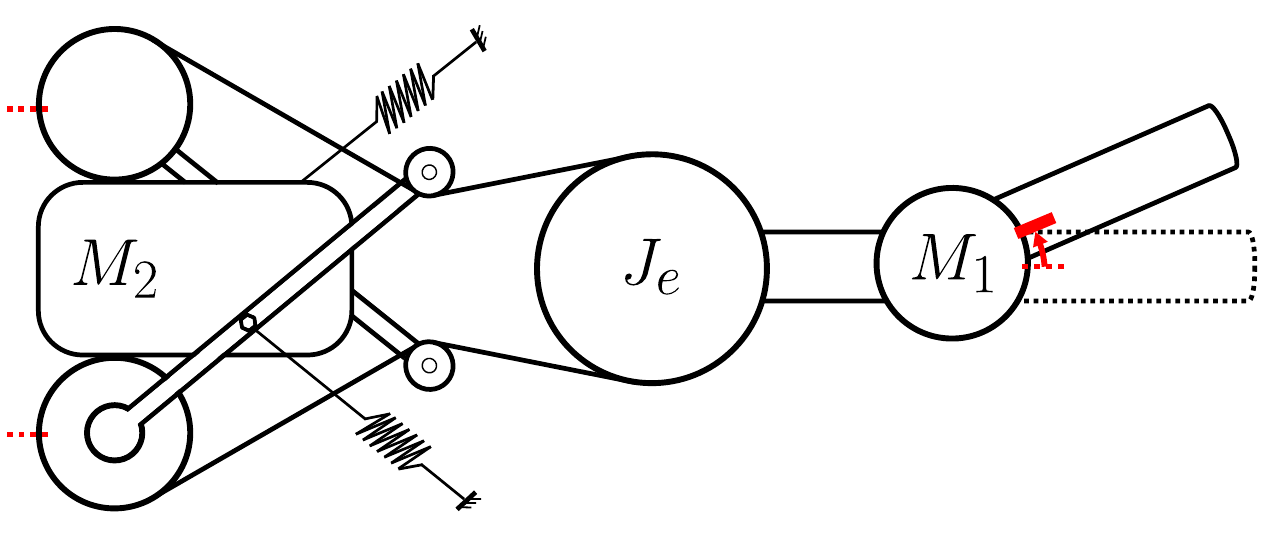}\label{fig:WorkingPrinciplesNew/3g}}
    \hfill    
    \subfloat[]{\includegraphics[width = 0.3\linewidth]{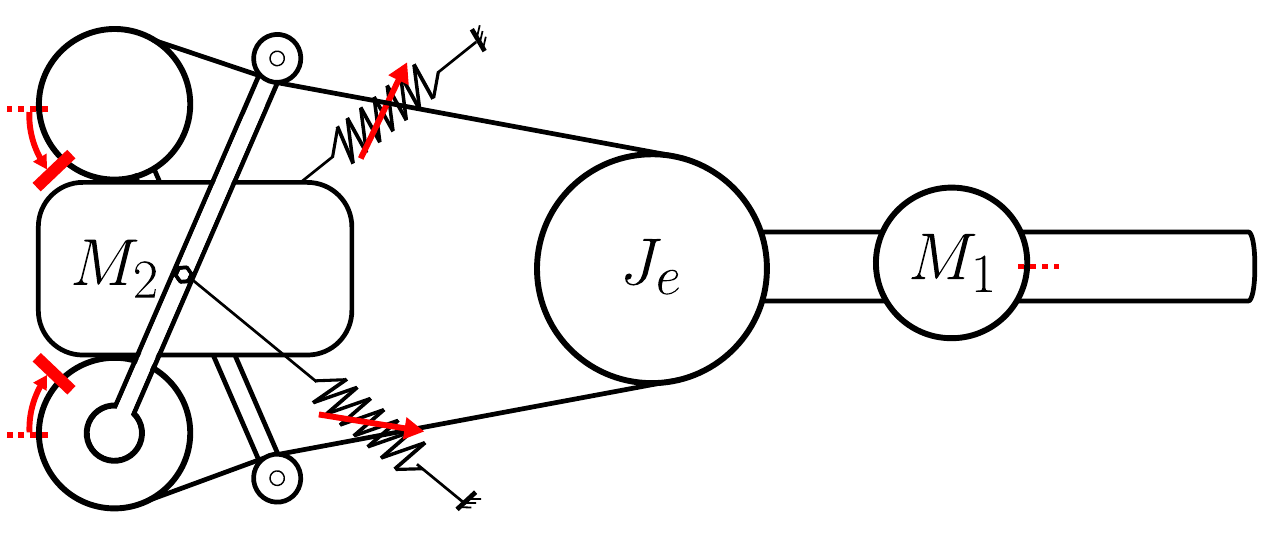}\label{fig:WorkingPrinciplesNew/3h}}
    \caption{Different implementations of the elastic transmission mechanism utilized in the VS-Elbow AA (a) and VS-Elbow D2 (b), and working principles of the VS-Elbow AA (c to e) and VS-Elbow D2 (f to h). To regulate the stiffness of the elastic transmission, the system modulates the active length of the belt (in blue), which adheres to the idle pulley (in red). As a result, the lever arm mechanism (in orange) rotates, changing the preload of linear extension springs (in green). Panels (c) and (f) show the system in neutral configuration. Panels (d) and (g) illustrate the regulation of the output position, while panels (e) and (h) display the stiffness modulation.}
    \label{fig:ElasticActuation}
\end{figure*}
The system exploits redundant actuation and a non-linear elastic transmission, shown in Fig.~\ref{fig:WorkingPrinciplesNew/3a} and \ref{fig:WorkingPrinciplesNew/3b}, to adjust joint stiffness by modulating the preload of extension springs.
Despite employing linear springs, non-linearity is achieved thanks to the geometry of the transmission, consisting of a belt stretched by a lever arm mechanism and idle pulleys.
Drawing inspiration from the human musculoskeletal system, two elastic mechanisms are strategically placed on opposite sides to exert an antagonistic effect on the elbow joint and ensure bidirectional joint stiffness. Consequently, in the absence of external loads, the net action of the two belts on the elbow shaft is null independently of the imposed spring preload.
When an external load is applied, the elastic joint deflects, and the elastic elements generate a restoring torque on the \mbox{F-segment} to achieve static equilibrium. 

\subsection{VS-Elbow AA}
The VS-Elbow AA employs two serial arrangements of a motor unit and elastic transmission, placed on opposite sides of the device to apply antagonistic torques (see Fig.~\ref{fig:WorkingPrinciplesNew/3c}). Therefore, one branch of the actuation system is analogous to the biceps brachii, while the other corresponds to the triceps brachii.
Within this setup, both motors are required to control either the equilibrium position or the stiffness of the joint. The output shaft accelerates when the net torque of the agonist branch exceeds that of the antagonist branch and the external load (Fig.~\ref{fig:WorkingPrinciplesNew/3d}). Additionally, the redundant actuation allows modulating the spring preload while maintaining zero net torque to regulate joint stiffness (Fig.~\ref{fig:WorkingPrinciplesNew/3e}).

\subsection{VS-Elbow D2}
The VS-Elbow D2 employs a VSA in the distributed independent configuration to achieve a human-like shape and mass distribution. This prototype is an optimized version of the device introduced in \cite{lemerle2019variable}, resulting in a lighter and more compact design. 

Considering the installation shown in Fig.~\ref{fig:2e}, the stiffness motor $M_2$ and the elastic transmission are located in the proximal part of the joint, while the position motor $M_1$ is fixed to the F-segment (see Fig.~\ref{fig:WorkingPrinciplesNew/3f}). 
To rotate the forearm, $M_1$ imparts a torque on the fixed elbow shaft, causing the F-segment to move around it (Fig.~\ref{fig:WorkingPrinciplesNew/3g}).
To adjust the joint stiffness, $M_2$ regulates the active length of both belts, modifying the preload of each spring and, consequently, the stiffness of the joint (Fig.~\ref{fig:WorkingPrinciplesNew/3h}).

%% file: Sections/Mechatronic.tex
\section{Mechatronic Design}\label{sect:Mechatronic Design}
\subsection{Mechanical Hardware}
\begin{figure*}[!ht]
    \centering
    \hspace{0.07\linewidth}
    \subfloat[]{\label{fig:VSE_AA}\includegraphics[height = 20 em]{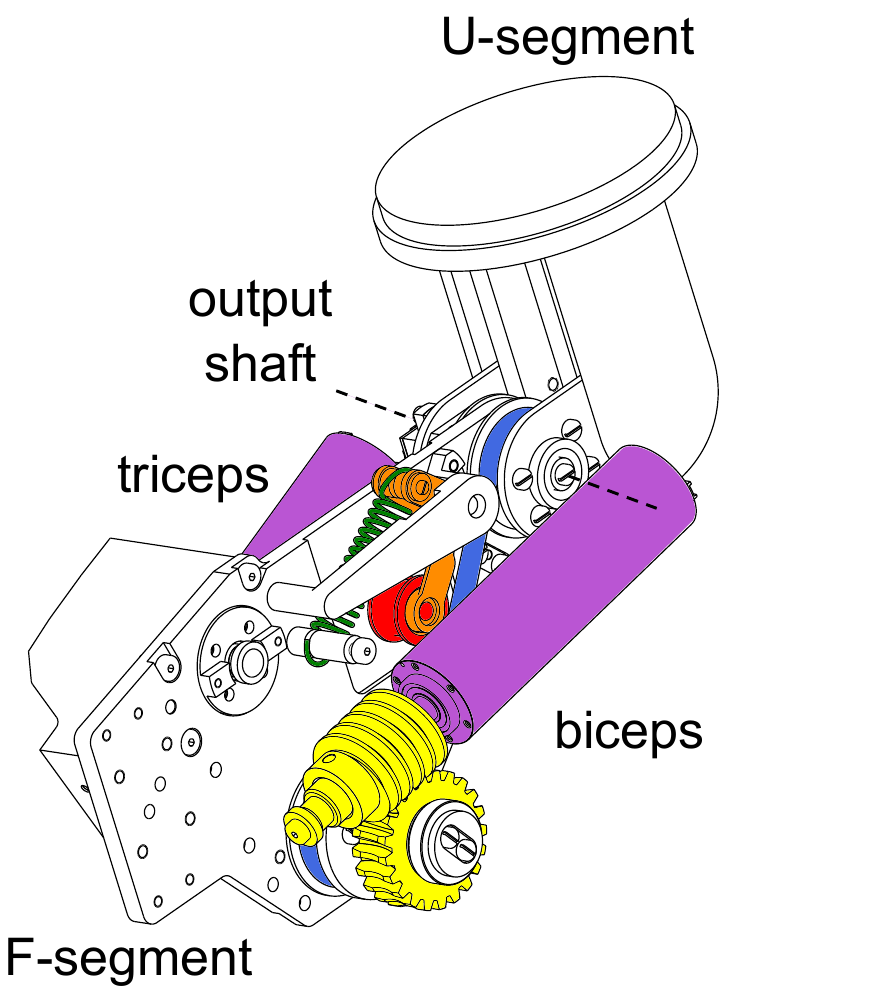}}
    \hfill
    \subfloat[]{\label{fig:VSE_D2}\includegraphics[height = 20 em]{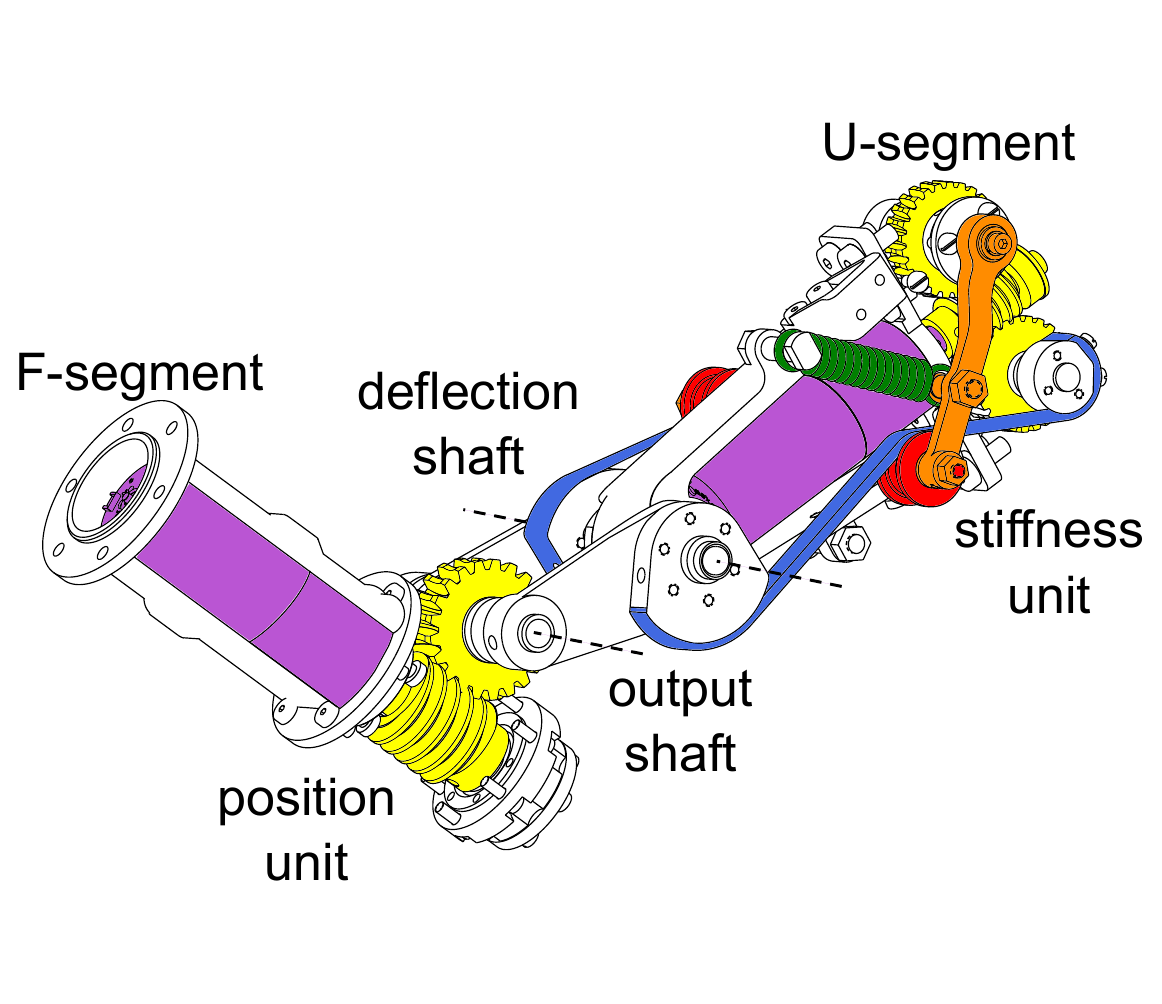}}
    \hspace{0.07\linewidth}
    \caption{CAD views of the two versions of the presented system. Panel (a) portrays the VS-Elbow AA, and panel (b) shows the VS-Elbow D2. The most significant components are highlighted in colors: yellow for worm drives, purple for DC motors, green for springs, red for idle pulleys, and orange for tensioning levers.}
    \label{fig:Cad4}
\end{figure*}

Fig.~\ref{fig:Cad4} displays the CAD of the presented system.
The elastic transmissions employed in the two versions of the VS-Elbow slightly differ in the geometric implementation of the lever arm mechanisms, although they adopt the same working principle outlined in Section~\ref{sect:work_princ}.
The actuation is conveyed to the elbow shaft by an open timing belt that wraps around the motor pulley, an idle pulley, and the elbow output shaft. The belt is secured to the motor and output shafts using screws.
The elastic elements are linear traction springs: T.091.110.0284.A from Vanel with a spring rate of 2.4 N/mm for the VS-Elbow AA, and AWF10-35 from Misumi with a spring rate of 4.17 N/mm for the VS-Elbow D2.

The motor transmission consists of the serial arrangement of a planetary gearbox and a low-efficiency worm drive, which ensures non-backdrivability. 
Non-backdrivable actuators provide significant advantages for prosthetic applications. The motors can utilize peak torques for active motion, while the worm gears maintain these torques passively and continuously, compensating for static loads and ensuring high joint stiffness without overloading the motors. Since prostheses are stationary most of the time, non-backdrivable actuation generally results in improved battery life and overall performance.
The maximum load the device can passively sustain thanks to the non-backdrivable actuation is limited by the tooth strength of the worm gears at 15.3 Nm. Table~\ref{tab:mech_elbow} summarizes the actuation components employed and their performance. 
The biceps motor is more powerful than the triceps since the first predominantly counteracts gravity. The asymmetrical motor choice saves space for fitting the output joint encoder and reduces the device size and weight. 
In the VS-Elbow AA, a worm screw rotates integrally with each motor output shaft and engages a worm gear, driving the motor output pulley. In the VS-Elbow D2, the worm screw of the stiffness motor unit engages two worm gears in phase opposition, actuating two output pulleys placed on opposite sides of the actuation system. In the position motor unit, instead, the worm gear is secured to the elbow shaft, preventing it from turning. As a result, the motor rotation causes the worm screw to revolve around the fixed gear, inducing the rotation of the F-segment relative to the U-segment.
Differently from the previous implementation, the axis of deflection and forearm rotation are not aligned but run parallel and are spaced by 40 mm to reduce the overall width of the positioning mechanism.
Compared to the device described in \cite{lemerle2019variable}, the VS-Elbow D2 results in being 46 mm thinner (-42\%), 33 mm shorter (-11\%), 849 g lighter (-53\%), without significant deterioration of performance. The fixed frame is made of aluminum alloy, while the cover and support parts are 3D-printed in ABS.
\begin{table}[!ht]
\small\sf\centering
\caption{Actuation components and performance. In the table, $\tau_n$ stands for the nominal torque, $\tau_m$ is the maximum active torque, $\xi$ is the transmission ratio, and $\eta$ is its efficiency.}
\label{tab:mech_elbow}
\renewcommand{\arraystretch}{1.5}
\resizebox{\columnwidth}{!}{%
\begin{tabular}{c|c|c|c|c}
\hline
\textbf{Elbow} & \textbf{Subsystem} & \textbf{Component} & \textbf{Model} & \textbf{Performance} \\
\hline
\multirow{7}{*}{AA} & \multirow{4}{*}{Biceps}    & Motor   & Maxon DCX22L GB KL 18V & \multirow{4}{*}{\begin{tabular}[c]{@{}c@{}} $\tau_n = 3.9$, $\tau_m = 8.7$ \\ $\xi = 336$, $\eta = 0.32$ \end{tabular}}  \\
                    &                            & Gearbox & Maxon GPX22UP 16:1 &                       \\
                    &                            & Worm    & FramoMorat A22U21 &                       \\
                    &                            & Springs & Vanel T.091.110.0284.A &                       \\ \cline{2-5} 
                    & \multirow{3}{*}{Triceps}   & Motor   & Maxon DCX22S GB KL 24V & \multirow{3}{*}{\begin{tabular}[c]{@{}c@{}} $\tau_n = 0.6$, $\tau_m = 3.9$ \\ $\xi = 111$, $\eta = 0.33\%$ \end{tabular}} \\
                    &                            & Gearbox & Maxon GPX22UP 5.3:1 &                       \\
                    &                            & Worm    & FramoMorat A22U21 &                       \\ \hline
\multirow{7}{*}{D2} & \multirow{4}{*}{Stiffness} & Motor   & Maxon DCX22S GB KL 24V & \multirow{4}{*}{\begin{tabular}[c]{@{}c@{}} $\tau_n = 2.4$, $\tau_m = 8.3$ \\ $\xi = 525$, $\eta = 0.30$ \end{tabular}} \\
                    &                            & Gearbox & Maxon GPX22 A 21:1 &                       \\
                    &                            & Worm    & FramoMorat A17U25 25:1 &                       \\
                    &                            & Springs & Misumi AWF10-35 &                       \\ \cline{2-5} 
                    & \multirow{3}{*}{Position}  & Motor   & DCX22L GB SL 18V & \multirow{3}{*}{\begin{tabular}[c]{@{}c@{}} $\tau_n = 5.5$, $\tau_m = 9.7$ \\ $\xi = 588$, $\eta = 0.27$ \end{tabular}} \\
                    &                            & Gearbox & Maxon GPX22 C 28:1 &                       \\
                    &                            & Worm    & FramoMorat A22U21 &                       \\ \hline
\end{tabular}%
}
\end{table}

\subsection{Electronic Hardware}
The device electronics and firmware architecture are based on the latest version of the Natural Machine Motion Initiative (NMMI) framework\footnote{\label{note3}\href{https://github.com/NMMI}{Natural Machine Motion Initiative}}.
The logic boards are PSoC 5LP CY8C58LP from Cypress and embed on a single chip configurable analog and digital peripherals, memory, and a microcontroller 32-bit Arm Cortex-M3. Although each logic board can control up to two motor drivers, we utilize two separate boards to command the two motors of the VS-Elbow D2 to optimize cable routing.
Two independent PID controllers regulate the position of the DC motors at a frequency of 1 kHz, relying on feedback from two rotational encoders.
Three 12-bit programmable rotary magnetic encoders AS5045 from Austriamicrosystems provide angular position feedback with a resolution of 0.04°. Among these, two encoders measure the rotation of the motor pulleys ($\theta_1$ and $\theta_2$), while the third one senses the deflection $\delta$ of the elastic joint or the position of the elbow shaft $\theta_o$ in the D2 and AA versions, respectively. Thus, the first two encoders provide feedback for the PID motor controllers, while the latter is necessary to determine the absolute position of the forearm.
The motor drivers, MC33887 from NXP Semiconductors, regulate the current flowing to the motor by modulating the duty cycle of a 20 kHz PWM signal. The maximum continuous current is 5 A, and the peak current is 7.8 A.
The logic boards are powered with 5 V through a USB connection to a PC, while both motors are powered by a single 24 V external battery. Subsequently, a firmware implementation of PWM rescaling enables the proper powering of the 18 V motor using a single power supply. The design and integration of the power supply system into the prosthesis are planned for future work. In the VS-Elbow AA, the battery can be placed either in the U-segment or between the F-segment and the wrist prosthesis. In the VS-Elbow D2, the battery should be positioned between the F-segment and the wrist or distributed along the volume of the device. To ensure maximum performance, the battery should be capable of delivering a peak current of 5 A to 7.6 A. Additionally, battery life should be evaluated considering also the specifications of the wrist and hand prostheses.

%% file: Sections/Control.tex
\section{Control}\label{sect:Control}
To control the position and stiffness of the elastic joint, we implemented two independent proportional control laws that regulate the position of each motor shaft using feedback from their respective rotational encoders.
Consequently, when an external load is applied to the elbow, the inherent elasticity allows a deflection between the U and F segments, causing the effective forearm position $\theta_o$ to deviate from the commanded one $\theta_p$.
Since the output position is measured with rotational encoders, we could potentially compensate for external loads by closing the feedback loop on the output position instead of the motor angles. However, this strategy would significantly alter the intrinsic compliance of the device by contrasting joint deflections, thereby increasing the joint stiffness and impeding its adaptive behavior \cite{keppler2018elastic}.

Due to its mechanical configuration, the VS-Elbow AA requires both motors to adjust either its stiffness or its position.
In the absence of external loads, the elbow joint angle $\theta_o$ is related to the motor angles $\theta_1$ and $\theta_2$ as 
\begin{equation}\label{eq:AA_pos_control}
    \theta_o = \frac{R_m}{2 R_o} (\theta_2 + \theta_1) = \frac{(\theta_2 + \theta_1)}{2} = \theta_p\;,
\end{equation}
since the radii of the motor and output pulleys $R_m$ and $R_o$ are equal.
To adjust the stiffness of the joint, we modulate the VS unit preload $\theta_s = \theta_2 - \theta_1$. 
Then, we control the two motor angles to independently regulate $\theta_p$ and $\theta_s$ holding
\begin{equation}\label{eq:AA_motor_control}
    \begin{pmatrix}
         \theta_{p}  \\
         \theta_{s} 
    \end{pmatrix}
     = 
    \begin{bmatrix}
        \frac{1}{2} & \frac{1}{2} \\
        -1 & 1
    \end{bmatrix}
     \begin{pmatrix}
          \theta_1  \\
          \theta_2
     \end{pmatrix}
     = A      
     \begin{pmatrix}
          \theta_1  \\
          \theta_2
     \end{pmatrix} \;.
\end{equation}
Since $A$ is full rank, the two control variables can be independently regulated via the motor inputs $\theta_1$ and $\theta_2$. 
Therefore, given a posture reference $\theta_p^{r}$ and a desired elastic preload $\theta_s^{r}$, inverting \eqref{eq:AA_motor_control} yields the reference motor angles as
\begin{equation}\label{eq:AA_ref}
    \begin{cases}
    \theta_1^{r} = \theta_p^{r} - \frac{\theta_s^{r}}{2}\\
    \theta_2^{r} = \theta_p^{r} + \frac{\theta_s^{r}}{2} 
\end{cases}\;.
\end{equation}

Controlling the VS-Elbow D2 is straightforward, as each motor unit independently regulates either the stiffness or the position of the joint, yielding $\theta_1^r = \theta_p^r$ and $\theta_2^r = \theta_s^r$.

\begin{figure*}[!t]
    \centering
    \includegraphics[width = \linewidth]{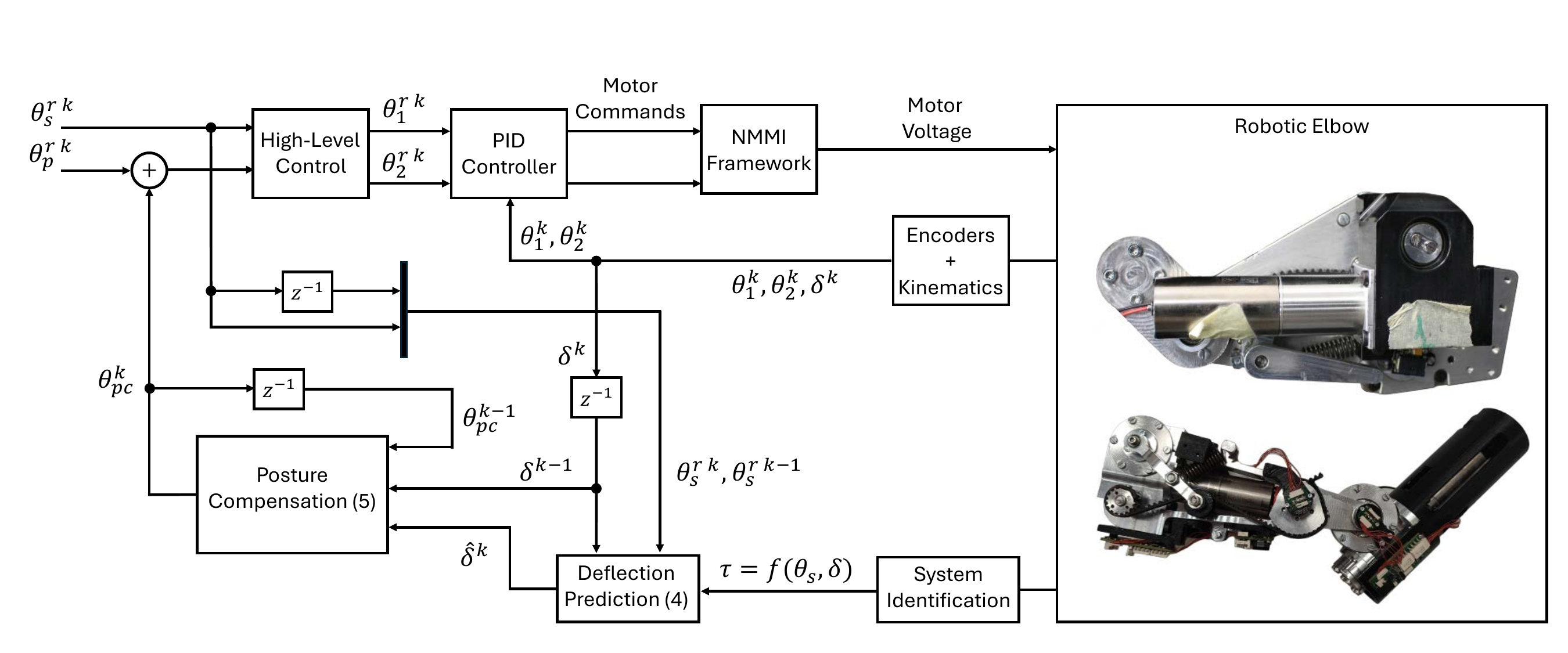}
    \caption{Block diagram schematizing the control strategy implemented.}
    \label{fig:PostureCompensation_Elbow}
\end{figure*}

\subsection{Posture Compensation of Stiffness Variations}
The proposed control strategy successfully achieves decoupled stiffness and position control in the absence of external loads.
However, if the user wishes to adjust the joint stiffness while an external load (e.g., the weight of the prosthesis) is applied, the elbow equilibrium position will change according to the new operating point of the non-linear spring. Motivated by usability considerations \cite{capsi2020exploring}, we developed a posture compensation strategy designed to minimize posture alterations resulting from stiffness variations.
The elastic torque of the VS joint, denoted as $\tau_s = f(\theta_s, \delta)$, is a non-linear function of the VS unit preload $\theta_s$ and deflection $\delta$.
To counteract the posture alteration that arises when changing the spring preload from $\theta_s^k$ to $\theta_s^{k+1}$, we enforce the elastic torque to remain constant during the transition, i.e., $\tau^k_s = \tau^{k+1}_s$. Assuming small variations between subsequent samples, we approximate $\tau_s^{k+1} = f(\theta_s^{k+1}, \delta^{k+1})$ with its first-order Taylor's approximation.  This assumption holds a reasonable approximation since the control loop bandwidth is much higher than the mechanical bandwidth of the system. Subsequently, imposing torque conservation yields the desired joint deflection as
\begin{equation} \label{eq:delta_comp}
    \hat{\delta}^{k+1} = \delta^k - \left.\left(\frac{\partial f^{-1}}{\partial \delta} \frac{\partial f}{\partial \theta_s}\right)\right|_{(\theta_s^{r\,k},\delta^k)}(\theta_s^{r\,k+1} - \theta_s^{r\,k}) \;,
\end{equation}
where $\partial f / \partial \delta$ represents the stiffness of the elastic joint, which is positive and invertible.  
Next, to ensure the conservation of the output shaft position $\theta_o^k = \theta_p^k + \delta^k = \theta_p^{k+1} + \hat{\delta}^{k+1} = \theta_o^{k+1}$, we iteratively compute the posture compensation command as 
\begin{equation}\label{eq:compensation_term}
\theta_{pc}^{k+1} = \theta_{pc}^k + \delta^k - \hat{\delta}^{k+1}\;. 
\end{equation}
Notably, $\partial f / \partial \theta_s$ is non-negative, indicating that an increase in the preload must correspond to a reduction in the deflection magnitude to maintain constant output torque. Furthermore, $\partial f / \partial \theta_s$ is zero in the unloaded configuration (i.e., $\delta^k = 0$). Therefore, in the absence of external loads, \eqref{eq:delta_comp} yields $\hat{\delta}^{k+1} = \delta^{k}$, meaning that the compensation term defined in \eqref{eq:compensation_term} does not introduce static errors in the posture tracking.
We did not employ this compensatory control while characterizing the VS joint to avoid altering the intrinsic properties of the device. Instead, we implemented this feature for active use during the case studies, adopting the torque function $f(\theta_s, \delta)$ identified during the characterization experiments, reported in Section \ref{sect:Experimental Protocol}.

\subsection{Hardware Implementation} \label{subsect:hardware implementation}
Fig.~\ref{fig:PostureCompensation_Elbow} schematize the control strategy implemented. The high-level controller and user interface are implemented in MATLAB Simulink using the NMMI libraries${}^{\ref{note3}}$, enabling real-time bidirectional communication between the Simulink environment and the electronics of the device. Due to the delay that arises from the communication protocol and graphic interface, the frequency of the Simulink diagram was lowered to 200 Hz to preserve real-time execution.

%% file: Sections/Experiments.tex
\section{Experimental Protocol}\label{sect:Experimental Protocol}
The experimental procedure employed in this study adheres to the guidelines outlined in \cite{grioli2015variable} for characterizing VSAs. The experimental protocol aims to evaluate the elastic properties, output link speed, and stiffness variation speed of the proposed system. Additionally, we present some case studies proving the benefits of adopting a VS prosthesis in unstructured environments, such as the real world.

\subsection{Elastic Joint Characterization}
\begin{figure}[!t]
    \centering
    \includegraphics[width = \linewidth]{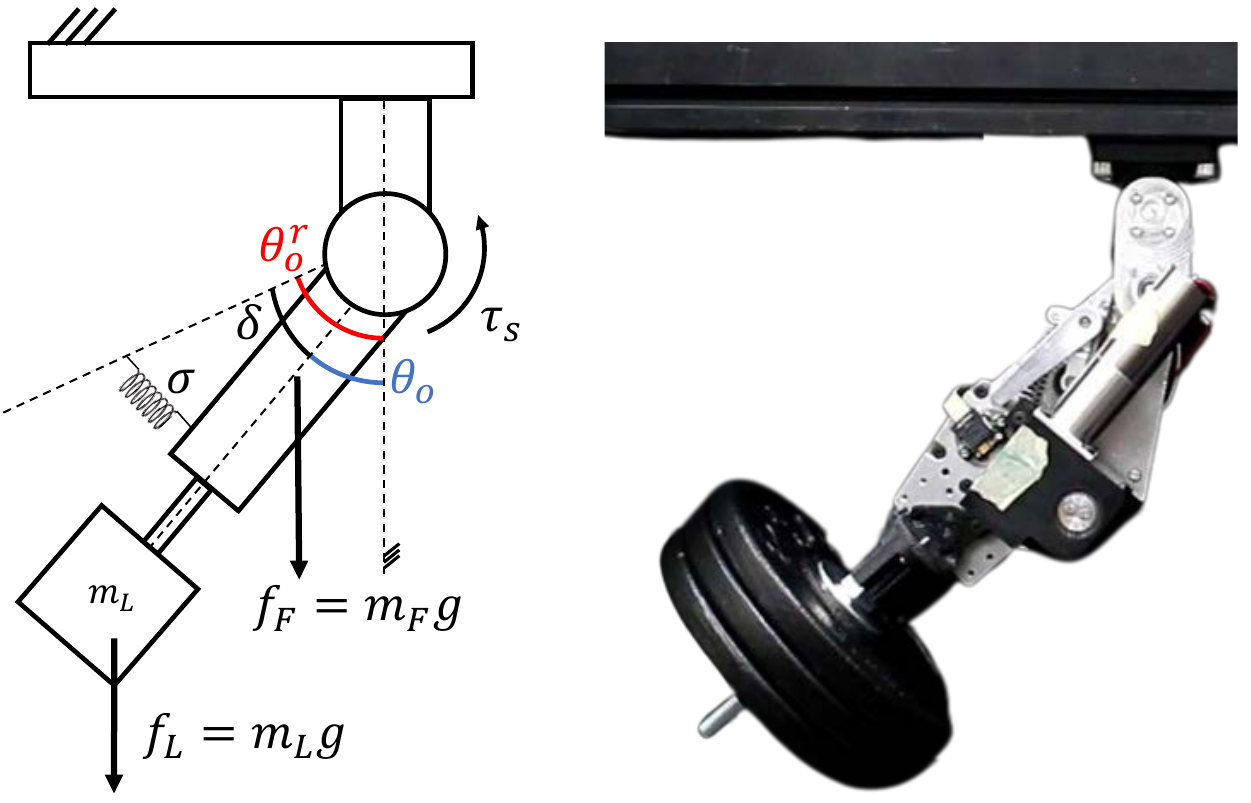}
    \caption{Experimental setup employed to characterize the elastic joint. A 3 kg load is attached to the end-effector, exerting a torque on the VS joint that varies with posture as the elbow explores its range of motion. On the left is a schematic representation of the VS-Elbow AA, and on the right is an image of the VS-Elbow AA during the experiments. The same setup was also used for the VS-Elbow D2.}
    \label{fig:Exp_setup}
\end{figure}

To characterize the elastic properties of the VS joint, we apply a known quasi-static external torque profile at various stiffness levels. 
The external torque is induced by attaching a known mass $m_L$ to the output link, as illustrated in Fig.~\ref{fig:Exp_setup}. 
Measuring $\theta_o$ with rotational encoders yields the torque acting on the elbow joint as 
\begin{equation}\label{eq: tau_sGravity}
    \tau(\theta_o) = \left(J_L^\top(\theta_o) m_L + J_{F}^\top(\theta_o) m_{F}\right)g\;,
\end{equation}
where the first component accounts for the external payload, and the second represents the contribution of the F-Segment mass $m_{F}$ applied in its center of gravity $F$. Here, $g$ is the gravity acceleration, and $J_*$ denotes the elbow Jacobian computed at the force application point $*$.

We characterize the system within its operating range, bounded by the peak motor torques.
By selecting an appropriate external load and safety margin, inverting \eqref{eq: tau_sGravity} provides the range of output positions $\overline{\theta}_o$ to explore during the experiments. Due to the varying joint deflection at different stiffness levels, the range of reference positions $\theta_p^r$ was adjusted for each stiffness configuration to achieve the desired output position $\overline{\theta}_o$. This range was covered through three repetitions of quasi-static sine waves with periods of 20 seconds and a payload of 3 kg.

\subsection{Output Link Speed Characterization}
The output link speed is evaluated using step references that span the entire RoM at different stiffness levels.
While the characteristics of the DC motors determine the joint speed as a function of the resisting torque, our interest extends to quantifying the relationship between the joint speed and stiffness.
During the experiment, to cancel the effect of the output load, the rotation axis of the VS joint is aligned with gravity. The output link velocity is computed by numerically differentiating the position measurements obtained from the encoders.

\subsection{Stiffness Variation Speed Characterization}
To assess the time required for changing joint stiffness, we modulate the VS unit preload from its minimum to maximum value using step commands, both with and without a 3 kg payload. The output joint position is set to reach the torque limit at maximum stiffness. 

\subsection{Case Studies}
We demonstrate that VS elastic actuation enables prostheses to safely interact with the environment and adapt to diverse, unpredictable situations by experimentally reproducing some real-world scenarios. The results for each scenario are reported for either one of the architectures to streamline the presentation.
A downloadable video demonstration is available as supplemental material to this paper. 

To show the effects of diverse joint stiffness on the system response to shocks, we hit the device with a hammer in minimum and maximum stiffness configurations. We then demonstrate how the system fosters safe interactions in unstructured environments. The transhumeral prosthesis impacts an obstacle while tracking a predefined posture reference in both low and high stiffness configurations. Interaction torques resulting from environmental constraints are estimated using the identified system characteristics and encoder measurements. Additionally, we assess the effect of low and high joint stiffness on payload capacity using a 2 kg external weight.
Finally, we implemented a biomimetic VS EMG controller inspired to \cite{capsi2020exploring} to demonstrate the prosthetic application. The interface consists solely of two commercial surface EMG electrodes 13E200 from Ottobock placed on the biceps and triceps brachii of a healthy subject. The controller regulates the joint speed, actuating elbow flexion and extension proportionally to the biceps and triceps brachii EMG activity while modulating joint stiffness proportionally to their coactivation.
\begin{figure*}[!t]
    \centering
    \subfloat[]{\includegraphics[width = 0.45\linewidth]{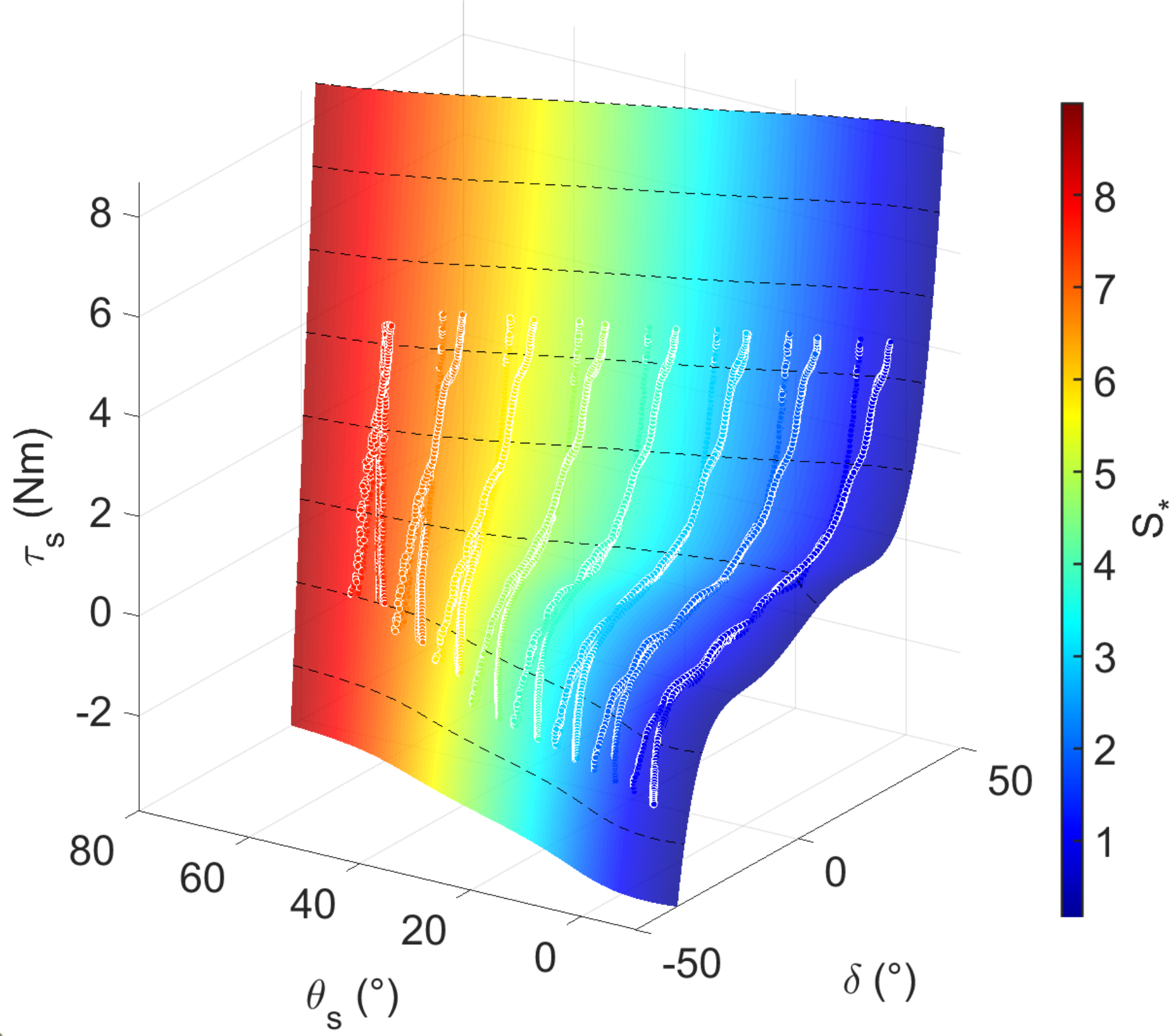}}
    \hfill
    \subfloat[]{\includegraphics[width = 0.45\linewidth]{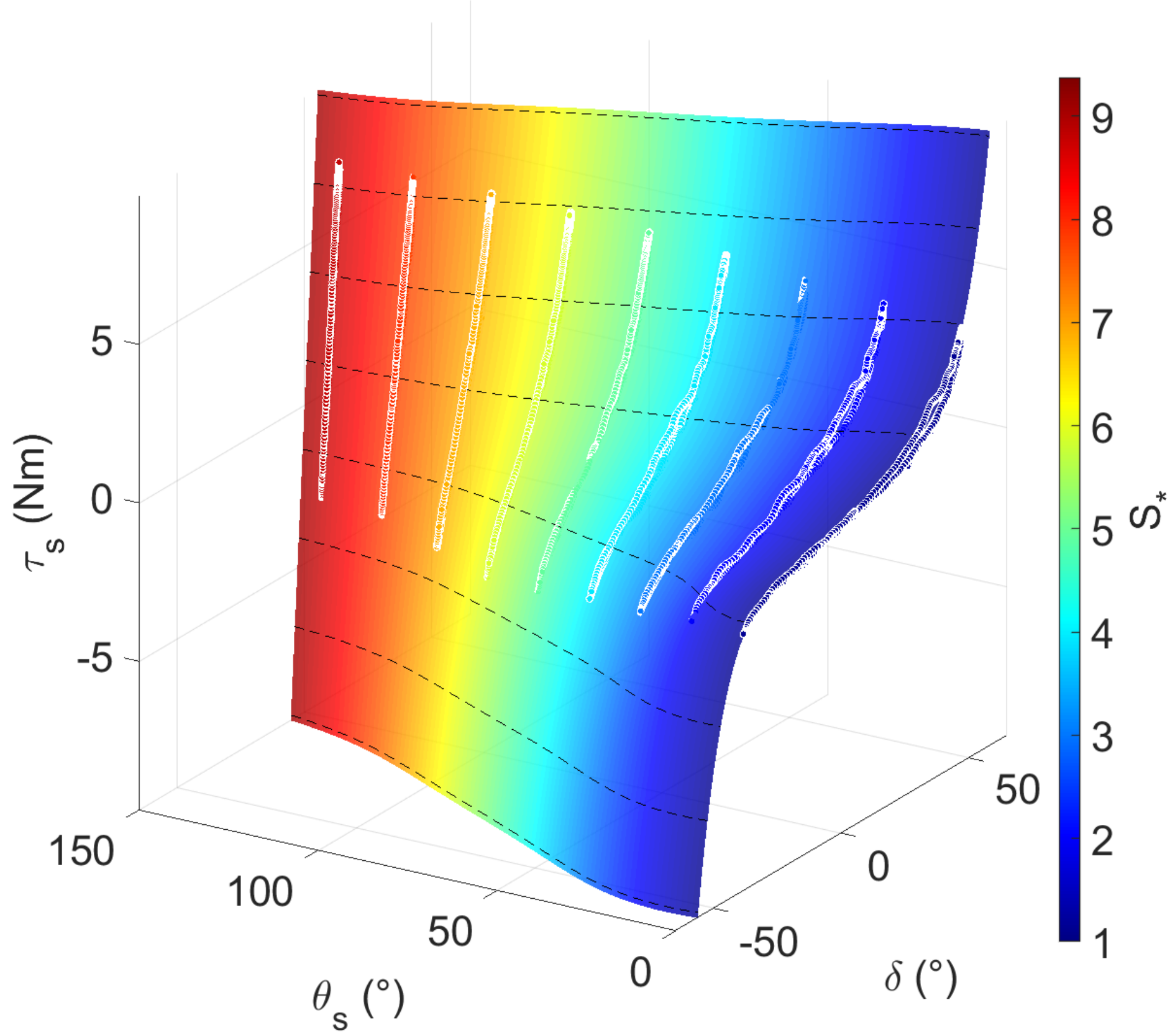}}
    \\
    \subfloat[]{\includegraphics[width = 0.45\linewidth]{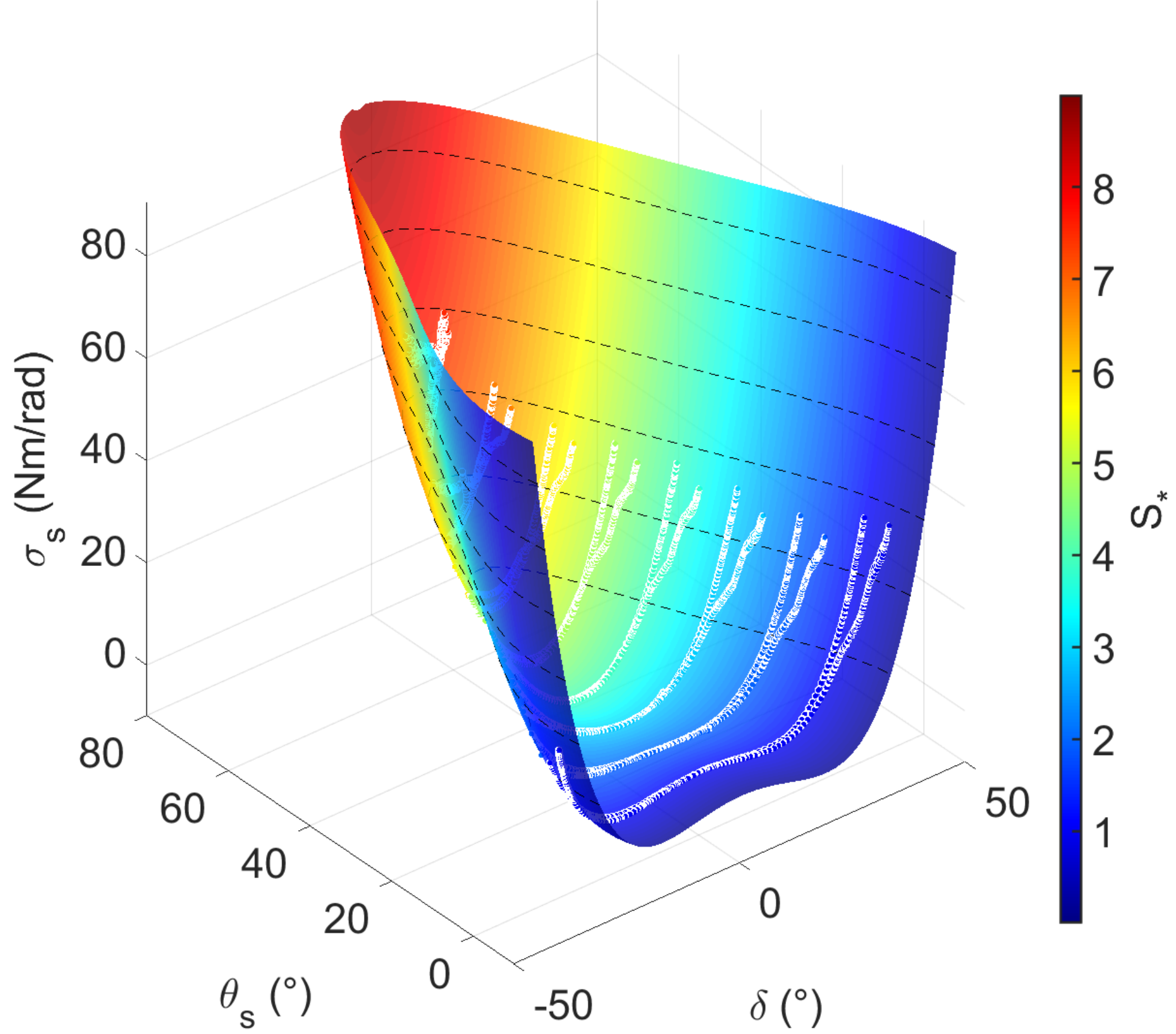}}
    \hfill
    \subfloat[]{\includegraphics[width = 0.45\linewidth]{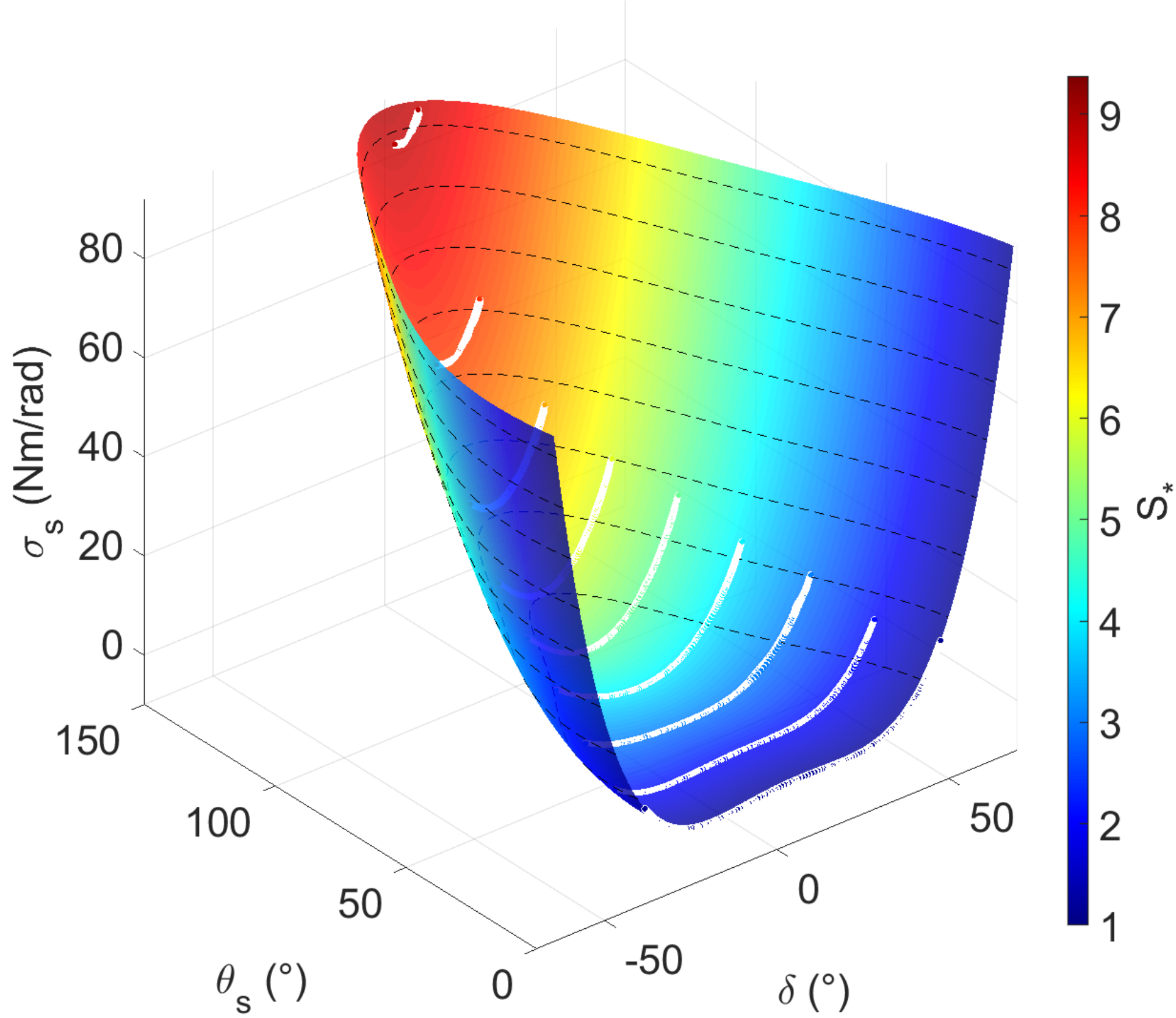}}
    \caption{Elastic joint characterization of the VS-Elbow AA (a and c) and VS-Elbow D2 (b and d). The sample data (white circles) in panels (a) and (b) represent the experimental output torque acting on the elastic joint at constant stiffness presets as the deflection varies. The color of the fitted polynomial surface is proportional to the VS unit preload, while the color bar indicates the tested stiffness configurations. The plots displayed in panels (c) and (d) represent the joint stiffness. The fitted polynomial surfaces are shown for the entire operating range of the system, bounded by peak motor torques.}
    \label{fig:Elastic_Exp}
\end{figure*}
\section{Results}\label{sect:results}
\begin{figure}[!t]
    \centering
    \subfloat[]{\includegraphics[height = 12.5 em]{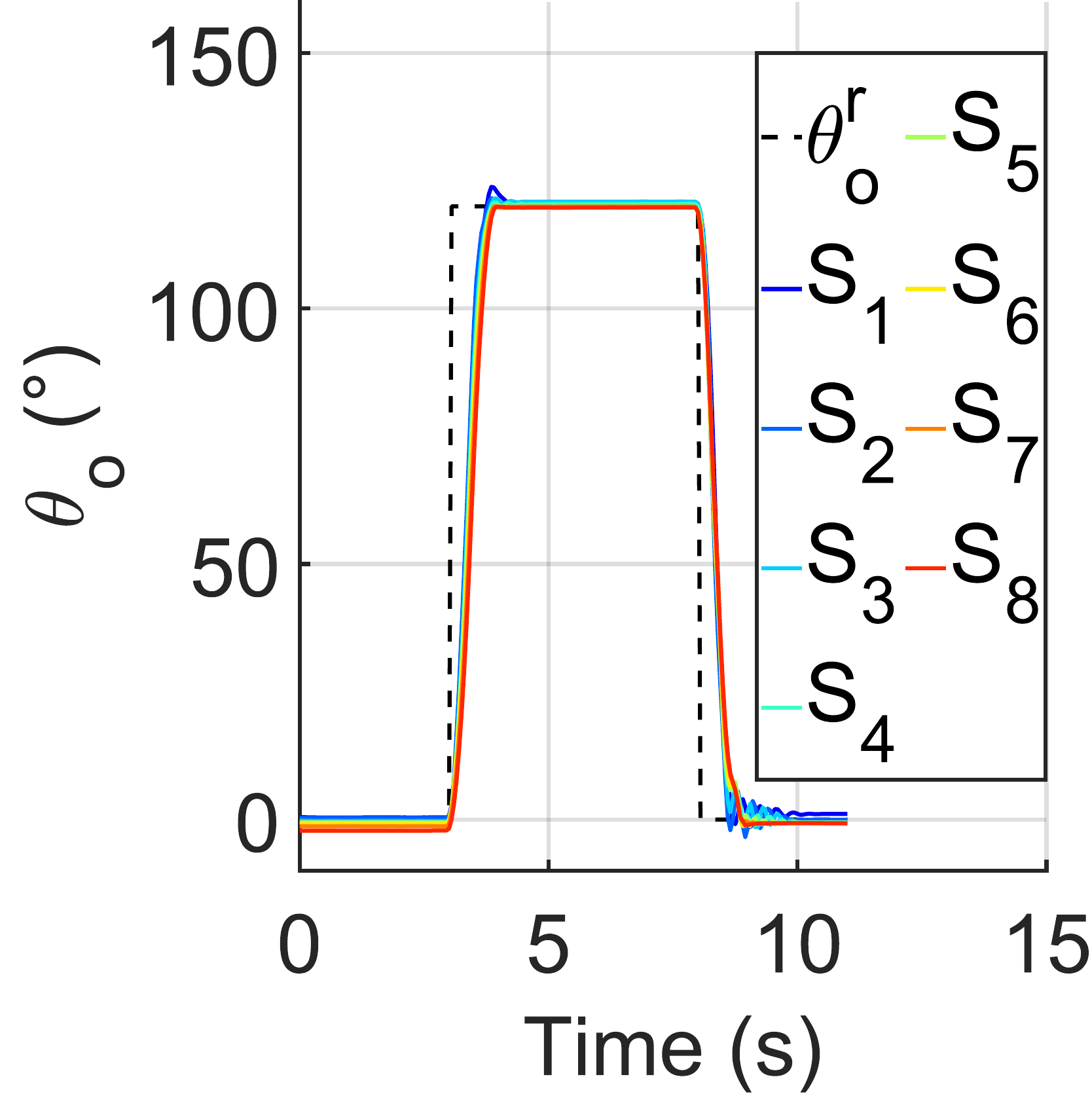}\label{fig:Position_Step_AA}}
    \hfill
    \subfloat[]{\includegraphics[height = 12.5 em]{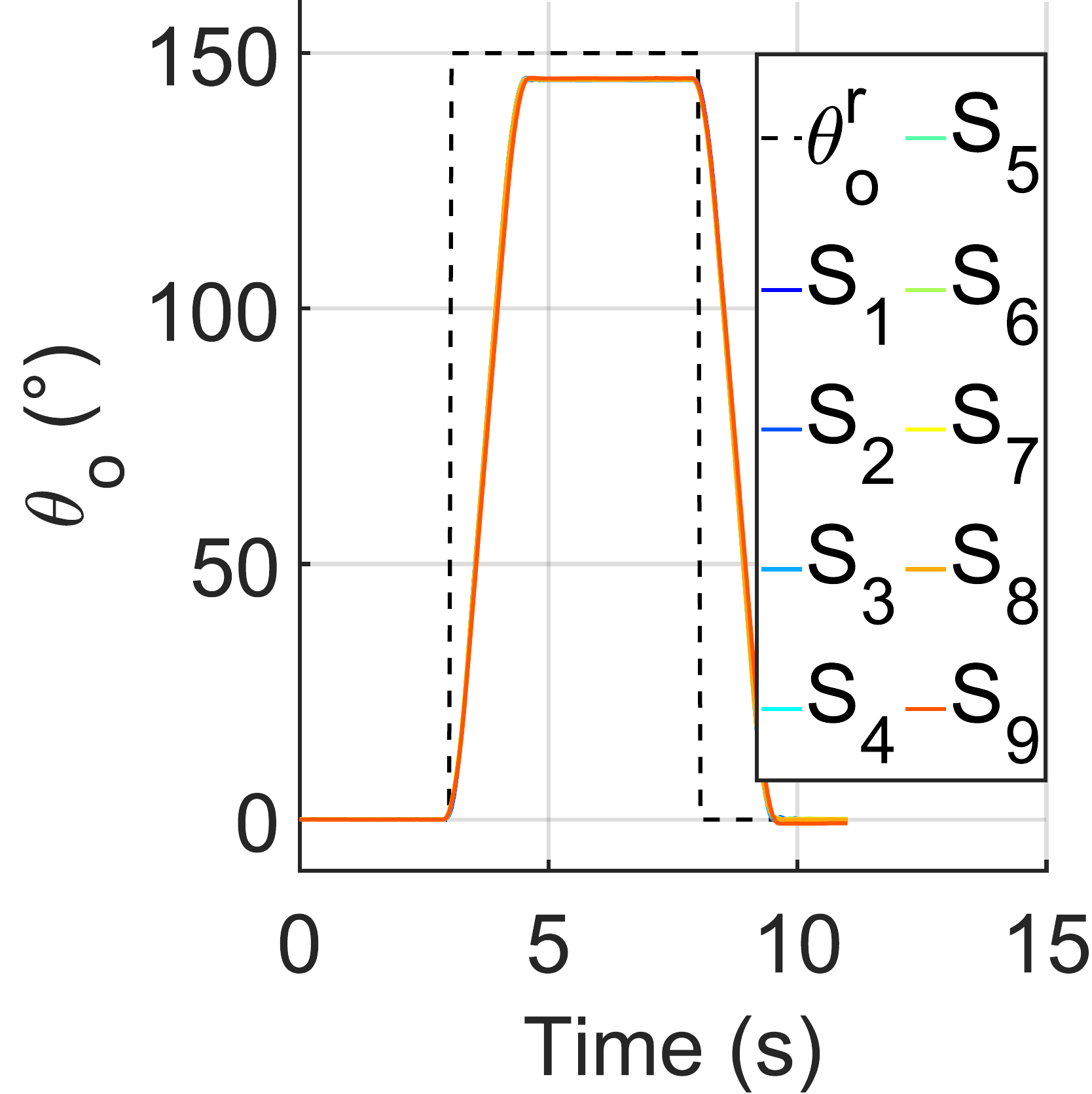}\label{fig:Position_Step_D2}}
    \caption{Output link position of the VS-Elbow AA (a) and D2 (b) in response to a step reference spanning the RoM of the device at different stiffness levels (increasing from $S_1$ to $S_9$).}
    \label{fig:Position_Step}
\end{figure}
\begin{figure}[!t]
    \centering
    \subfloat[]{\includegraphics[height = 12.9 em]{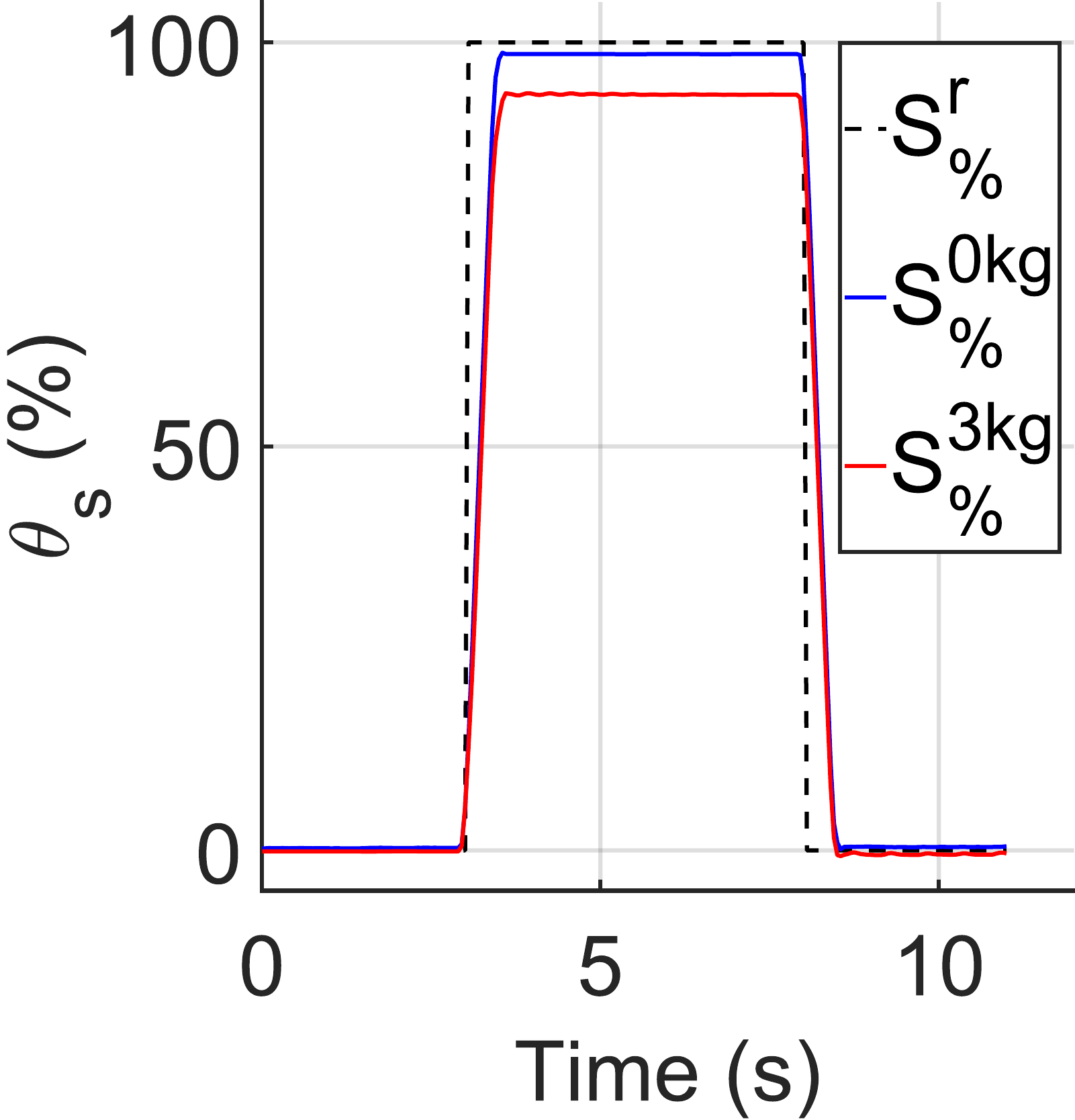}\label{fig:Stiffness_Step_AA}}
    \hfill
    \subfloat[]{\includegraphics[height = 12.9 em]{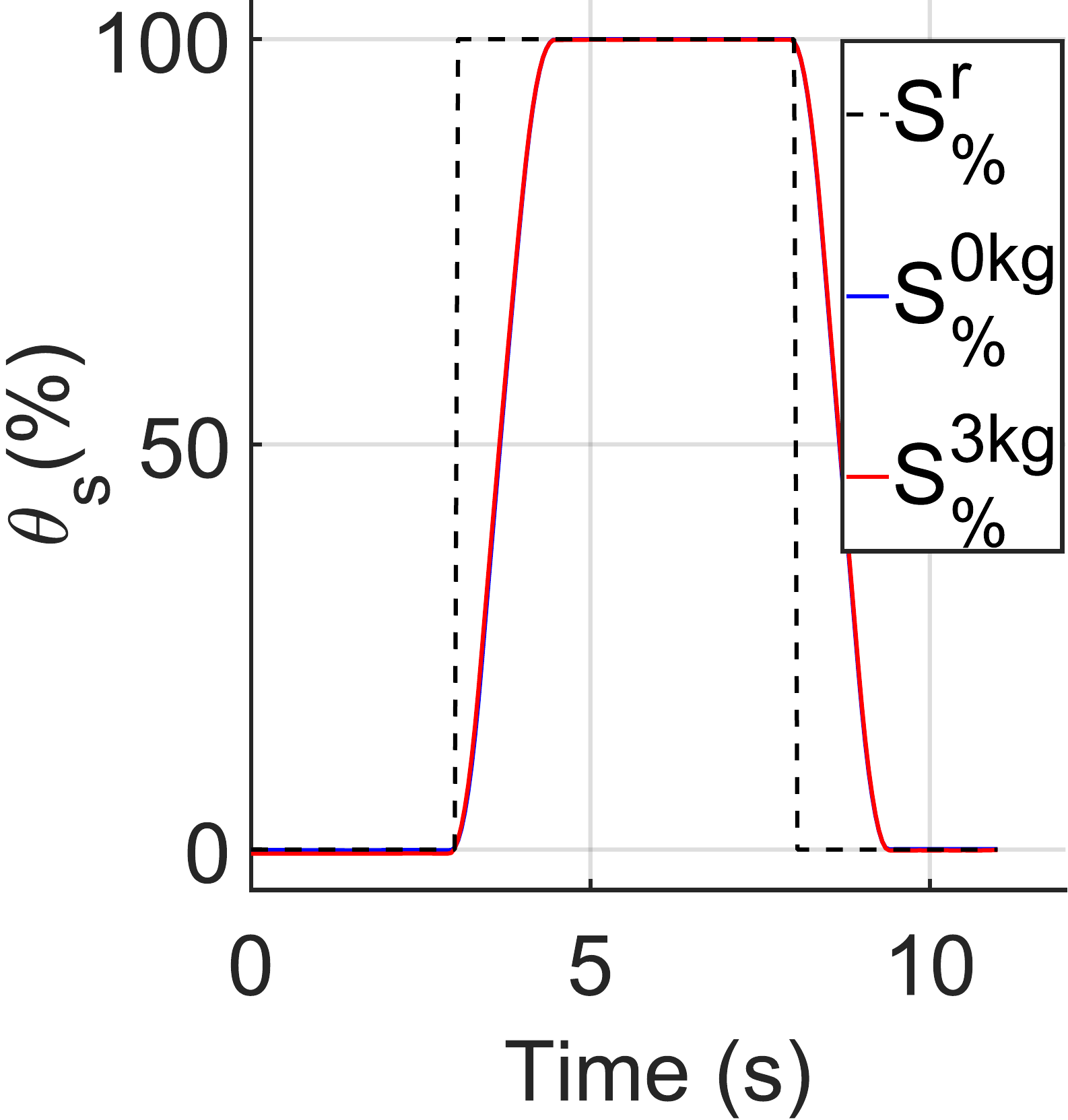}\label{fig:Stiffness_Step_D2}}
    \caption{Stiffness preload of the VS-Elbow AA (a) and D2 (b) in response to a step reference varying from the minimum to the maximum stiffness, in unloaded condition (blue lines) and with a payload of 3 kg (red lines).}
    \label{fig:Stiffness_Step}
\end{figure}
The complete system characterization can be found in the datasheets attached as supplemental material to this paper. The most salient results are reported in this section and summarized in Table~\ref{tab:comp_elbow}.

\subsection{Elastic Joint Characterization}
Fig.~\ref{fig:Elastic_Exp} illustrates the elastic behavior of the VS joints. The experimental data have been fitted with a 2D polynomial of the 5th order in joint deflection $\delta$ and 4th order in stiffness preload $\theta_s$, achieving coefficients of determination $R^2_{AA} = R^2_{D2} = 0.99$ and root mean square errors $RMSE_{AA} = 0.19$ Nm and $RMSE_{D2} = 0.27$ Nm. The polynomial coefficients are obtained through least square optimization and are detailed in the attached datasheets.
The output stiffness is derived by differentiating the 2D polynomial with respect to $\delta$, and the achievable range is reported in Table~\ref{tab:comp_elbow}. From the same data, we also extracted the maximum deflection $\Delta$ and hysteresis $H$ observed at minimum (m) and maximum (M) stiffness. For the VS-Elbow AA, we obtained $\Delta^{AA}_{m} = 43.9$°, $\Delta^{AA}_{M} = 7.9$°, $H^{AA}_{m} = 2.9$°, $H^{AA}_{M} = 1.4$, while the VS-Elbow D2 yielded $\Delta^{D2}_{m} = 47.6$°, $\Delta^{D2}_{M} = 5.1$°, $H^{D2}_{m} = 7$°, $H^{D2}_{M} = 0.9$°.

\subsection{Output Link Speed}
Fig.~\ref{fig:Position_Step} illustrates the position of the terminal link in response to a step reference exploring the entire device RoM at different stiffness levels. 
Fig.~\ref{fig:Position_Step_AA} demonstrates that the output speed of the VS-Elbow AA is influenced by the joint stiffness. Additionally, due to the use of different motors on each side, the elbow flexion speed is lower than the extension speed. The maximum speed is achieved at minimum stiffness, reaching 265 °/s for flexion and -306 °/s for extension, while at maximum stiffness, the highest speeds are 232 °/s and -238 °/s for flexion and extension, respectively.
In Fig.~\ref{fig:Position_Step_D2}, instead, it is evident that in the distributed implementation, the maximum speed remains constant at 127.5 °/s, regardless of the imposed spring preload or the direction of motion. 

\subsection{Stiffness Variation Speed}
Fig.~\ref{fig:Stiffness_Step} illustrates the stiffness step response of the device. Fig.~\ref{fig:Stiffness_Step_AA} demonstrates that for the AA architecture, the output load opposes the stiffness variation, increasing the stiffness variation time (SVT) and residual error. The time required to transition from 10\% to 90\% of the maximum spring preload, with and without load, is $SVT^{AA}(3 \, \text{kg}) = 0.6$ s and $SVT^{AA}(0 \, \text{kg}) = 0.5$ s. On the other hand, in Fig.~\ref{fig:Stiffness_Step_D2}, it is evident that for the VS-Elbow D2, the SVT is independent of the output load, remaining constant at $SVT^{D2} = 1.15$ s. 

\subsection{Case Studies}
Fig.~\ref{fig:Hammer} illustrates the system response to sudden impacts at low (Fig.~\ref{fig:Hammer_a}) and high (Fig.~\ref{fig:Hammer_b}) joint stiffness. Fig.~\ref{fig:Hammer_c} reports that at low joint stiffness, sudden impacts induce damped forearm oscillations around an average value of $\delta = 6$°, with a peak amplitude of $10$°, and a mean lifetime of $0.7$ s. 
Conversely, high joint stiffness eliminates oscillations and provokes impulsive deflections smaller than $1$° after the impacts. 

Fig.~\ref{fig:Obstacle} portrays the VS-Arm AA encountering a fixed obstacle while moving at low (Fig~\ref{fig:Obstacle_a}) and high (Fig~\ref{fig:Obstacle_b}) elbow stiffness. Fig.~\ref{fig:Obstacle_c} shows that the maximum elastic torque generated by the environmental constraint on the elbow shaft is significantly lower in the soft ($\tau_s = -1.9$ Nm) than in the rigid configuration ($\tau_s = -9.4$ Nm).

Fig.~\ref{fig:Bag_a} demonstrates that the VS-Elbow D2 is unable to sustain a load of 2 kg in the soft configuration. In contrast, it effectively supports the payload at high joint stiffness (Fig.~\ref{fig:Bag_b}), resulting in a maximum deflection of $1.5$°.

Fig.~\ref{fig:EMG_control} showcases the EMG control application. The algorithm correctly maps the user's intention within the bionic limb. However, small undesired movements may occur before detecting a pure cocontraction, owing to the imperfect synchronization of muscular activity on either EMG channel.

%% file: Sections/Discussions.tex
\section{Discussions}\label{sec:discussions}
\begin{table*}
\small\sf\centering
\caption{Comparative table of morphological and functional characteristics of human and robotic prosthetic elbows. 
The torque reported for the prostheses is intended as peak torque, which cannot be maintained continuously with backdrivable actuation. Additionally, the maximum passive load is specified for non-backdrivable mechanisms. A rough estimate of the maximum load referred to the hand can be obtained as $W \approx \tau / 3$ (in kg). For VS devices, it is reported whether the stiffness is modified passively by modulating the transmission physical properties or actively through software control.}
\label{tab:comp_elbow}
\renewcommand{\arraystretch}{1.5}
\begin{tabular*}{\textwidth}{@{\extracolsep{\fill}} c c c c c c c c c c}
\hline
\textbf{Name} & \textbf{Segment} & \textbf{D (mm)} & \textbf{L (mm)} & \textbf{m (g)} & \textbf{RoM (°)} & $\mathbf{\tau}$ \textbf{(Nm)} & \textbf{$\mathbf{\omega}$ (°/s)} & \textbf{$\mathbf{\sigma}$ (Nm/rad)} \\
\hline
\begin{tabular}{@{}c@{}}Median Male \\ Elbow Segments \end{tabular} & \begin{tabular}{@{}c@{}} U \\ F \\\end{tabular} & \begin{tabular}{@{}c@{}} 99 \\ 98 \\\end{tabular} & \begin{tabular}{@{}c@{}} 183 \\ 135 \\\end{tabular} & \begin{tabular}{@{}c@{}} 1250 \\ 725 \\\end{tabular} & [0, 120]$^\ast$ & [-2.8, 5.9]${}^\ast$ & $\leq$ 250${}^\ast$ & [2, 60]${}^{\ast,} {}^\text{p}$\\
\hline
\begin{tabular}{@{}c@{}}Median Female \\ Elbow Segments \end{tabular} & \begin{tabular}{@{}c@{}} U \\ F \\\end{tabular} & \begin{tabular}{@{}c@{}} 81 \\ 83 \\\end{tabular} & \begin{tabular}{@{}c@{}} 149 \\ 122 \\\end{tabular} & \begin{tabular}{@{}c@{}} 940 \\ 545 \\ \end{tabular} & [0, 120]$^\ast$ & [-2.8, 5.9]${}^\ast$ & $\leq$ 250${}^\ast$ & [2, 60]${}^{\ast,} {}^\text{p}$\\
\hline
VS-Elbow AA &  U/F & 85 & 133 & 894 & [0, 120] & [-3.9, 8.7]($\pm$15.3)${}^\dagger$ & [-306, 265] & [0.65, 77]${}^\text{p}$\\
\hline
VS-Elbow D2 & \begin{tabular}{@{}c@{}} U \\ F \\\end{tabular} & \begin{tabular}{@{}c@{}} 64 \\ 62 \\\end{tabular} & \begin{tabular}{@{}c@{}} 127 \\ 124 \\\end{tabular} & \begin{tabular}{@{}c@{}} 522 \\ 413 \\\end{tabular} & [0, 142] &$\pm$9.7($\pm$15.3)${}^\dagger$ & $\pm$128 & [1, 93]${}^\text{p}$ \\
\hline
VS-Elbow D \cite{lemerle2019variable} & \begin{tabular}{@{}c@{}} U \\ F \\\end{tabular} & \begin{tabular}{@{}c@{}} 110 \\ 65 \\\end{tabular} & \begin{tabular}{@{}c@{}} 160 \\ 120 \\\end{tabular} & \begin{tabular}{@{}c@{}} 955 \\ 820 \\\end{tabular} & [-25, 150] & $\pm$15($\pm$15.3)${}^\dagger$ & $\pm$101 & [1, 390]${}^\text{p}$\\
\hline
Baggetta et al. \cite{baggetta2022design} & \begin{tabular}{@{}c@{}} U \\ F \\\end{tabular} & \begin{tabular}{@{}c@{}} 95 \\ 92 \\\end{tabular} & \begin{tabular}{@{}c@{}} 165 \\ 156 \\\end{tabular} & \begin{tabular}{@{}c@{}} 920 \\ 419 \\\end{tabular} & [-100, 100] & - & - & [2, 16]${}^\text{p}$\\
\hline
Bennet et al. \cite{bennett2016design} & \begin{tabular}{@{}c@{}} U \\ F \\\end{tabular} & \begin{tabular}{@{}c@{}} - \\ 50 \\\end{tabular} & \begin{tabular}{@{}c@{}} 30 \\ 174 \\\end{tabular} & \begin{tabular}{@{}c@{}} 1000 \\\end{tabular} & [15, 145] & $\pm$18.4 & $\pm$490 & \begin{tabular}{@{}c@{}} rigid, \\ free-swing${}^\text{a}$ \end{tabular}\\
\hline
Sensinger et al. \cite{sensinger2008user} & F & 70 & 70 & 740 & - & $\pm$50 & $\pm$140 & [2, 102]${}^\text{a}$\\
\hline
RIC Elbow \cite{lenzi2016ric} & F &  76 & - & 746 & - & $\pm$12($\pm$68)${}^\dagger$ & $\pm$80 & fixed \\
\hline
\begin{tabular}{@{}c@{}} Fillauer \\ Utah U3+${}^\text{c,\,}$${}^{\ref{note1}}$ \\\end{tabular} & F & - & 270 & 913 & [20, 155] & $\pm$3.1($\pm$69)${}^\dagger$ & - & \begin{tabular}{@{}c@{}} rigid, \\ free-swing${}^\text{a}$ \end{tabular}\\
\hline
\begin{tabular}{@{}c@{}} Ottobock \\ Dynamic Arm${}^\text{c,\,}$${}^{\ref{note2}}$ \\\end{tabular} & F & - & 207 & 1000 & [15 145] & $\pm$16($\pm$70)${}^\dagger$ & [-, 260] & \begin{tabular}{@{}c@{}} rigid, \\ free-swing${}^\text{p}$ \end{tabular}\\
\hline
\multicolumn{9}{c}{\begin{tabular}{@{}c@{}}  \footnotesize{${}^\ast$ Functional for standard activities of daily living. ${}^\text{p}$ Passively regulated. ${}^\text{a}$ Actively regulated. ${}^\dagger$ Maximum passive load. ${}^\text{c}$ Commercial device.} \end{tabular}}\\
\end{tabular*}
\end{table*}
\begin{figure}
    \centering
    \subfloat[]{\includegraphics[width = 0.49 \linewidth]{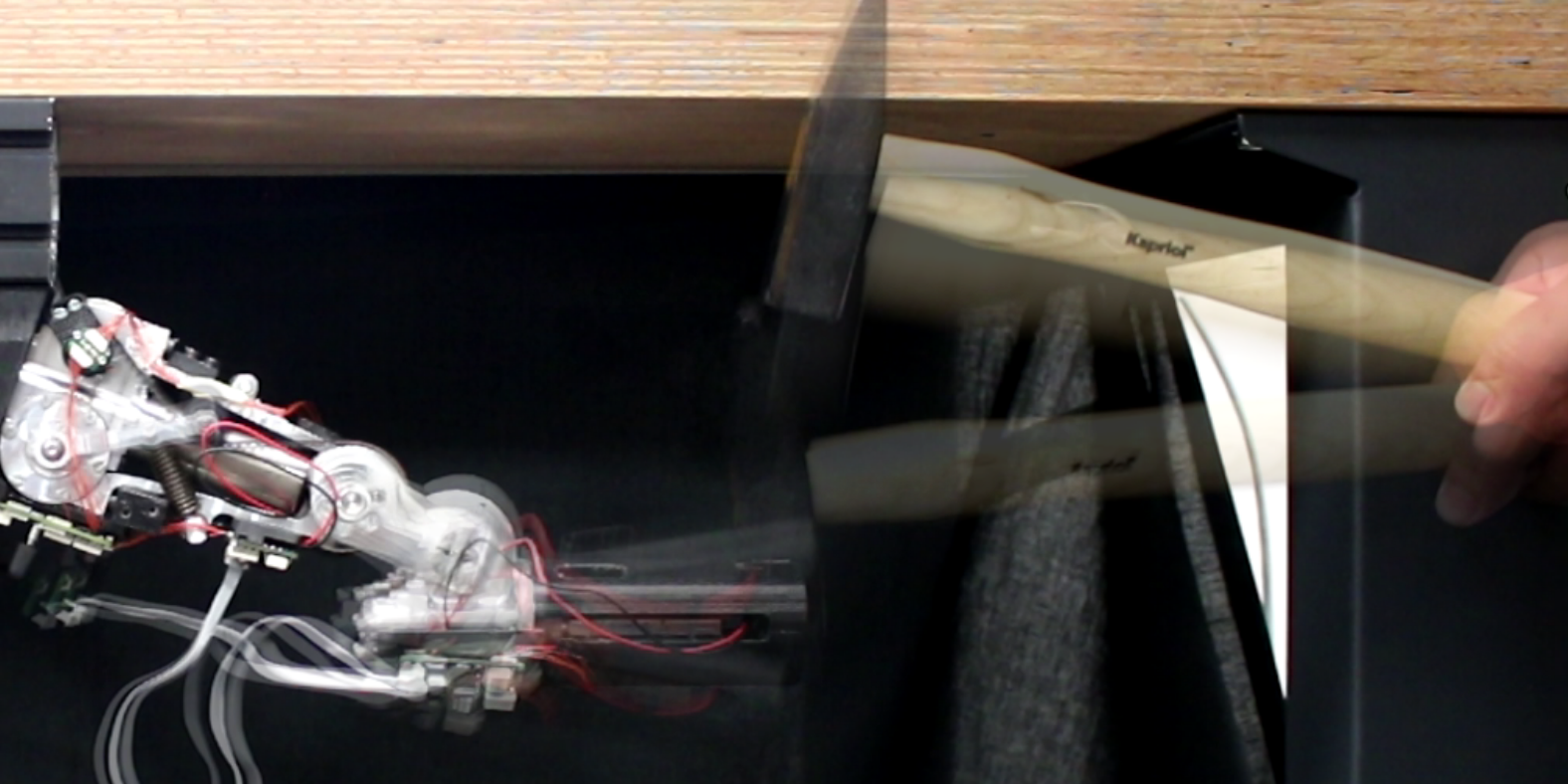}\label{fig:Hammer_a}}
    \hfill
   \subfloat[]{\includegraphics[width = 0.49 \linewidth]{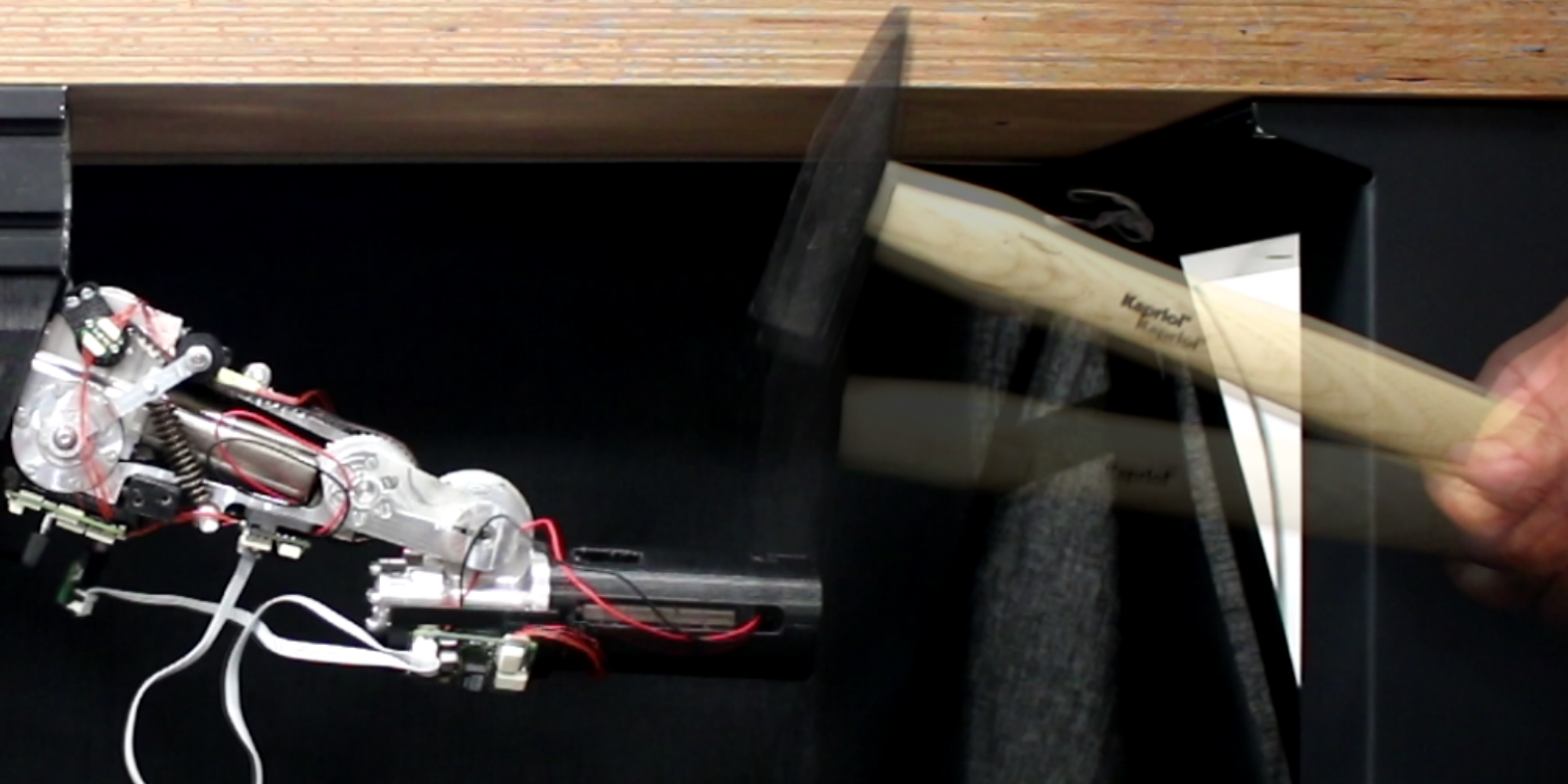}\label{fig:Hammer_b}} \\
    \subfloat[]{\includegraphics[width = 0.98 \linewidth]{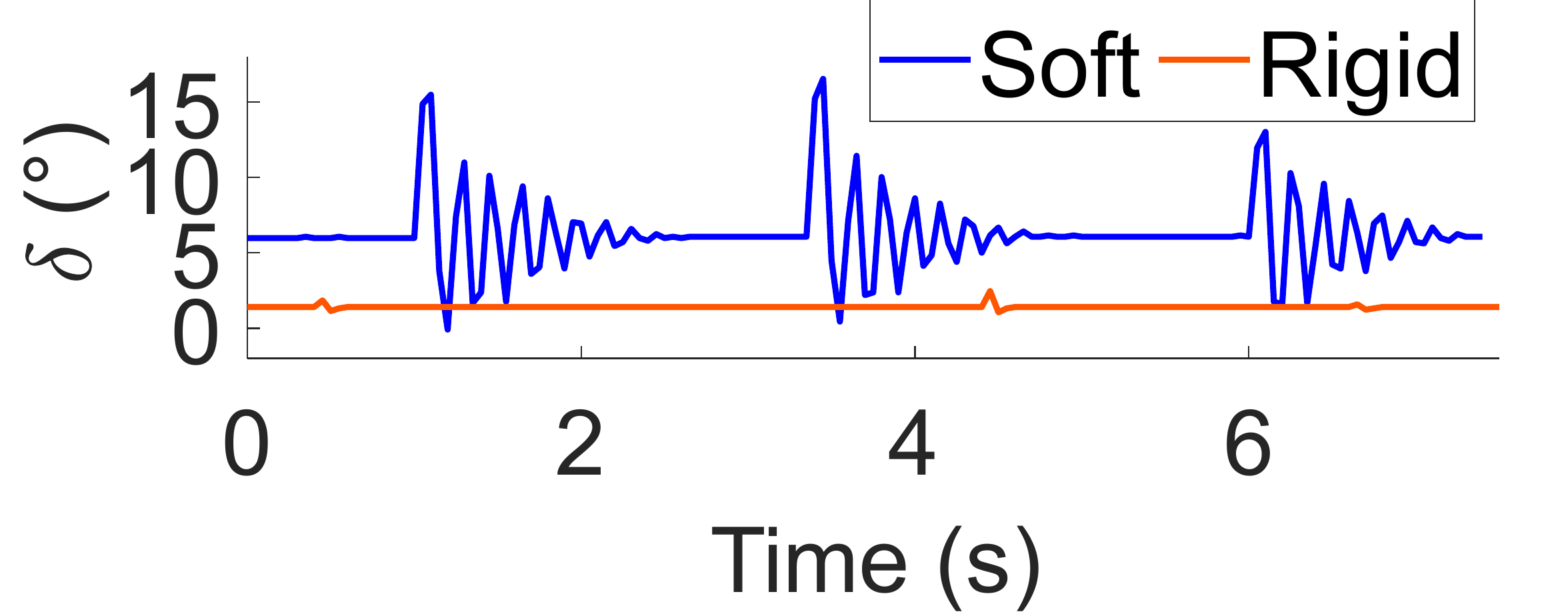}\label{fig:Hammer_c}}
    \caption{System response to sudden impacts: the VS-Elbow D2 is struck by a hammer in both soft (a) and rigid (b) configurations. The graph (c) depicts the deflection of the elastic joint after the impacts. In the soft configuration (blue line), the springs absorb the shocks, resulting in damped oscillations. Conversely, in the rigid configuration (orange line), the elbow resists the impacts, generating high, impulsive reaction torques that can pose a risk to the user and the device itself.  The static deflection is determined by the torque required to balance the weight of the F-segment.}
    \label{fig:Hammer}
\end{figure}
\begin{figure}
    \centering
    \subfloat[]{\includegraphics[width = 0.49\linewidth]{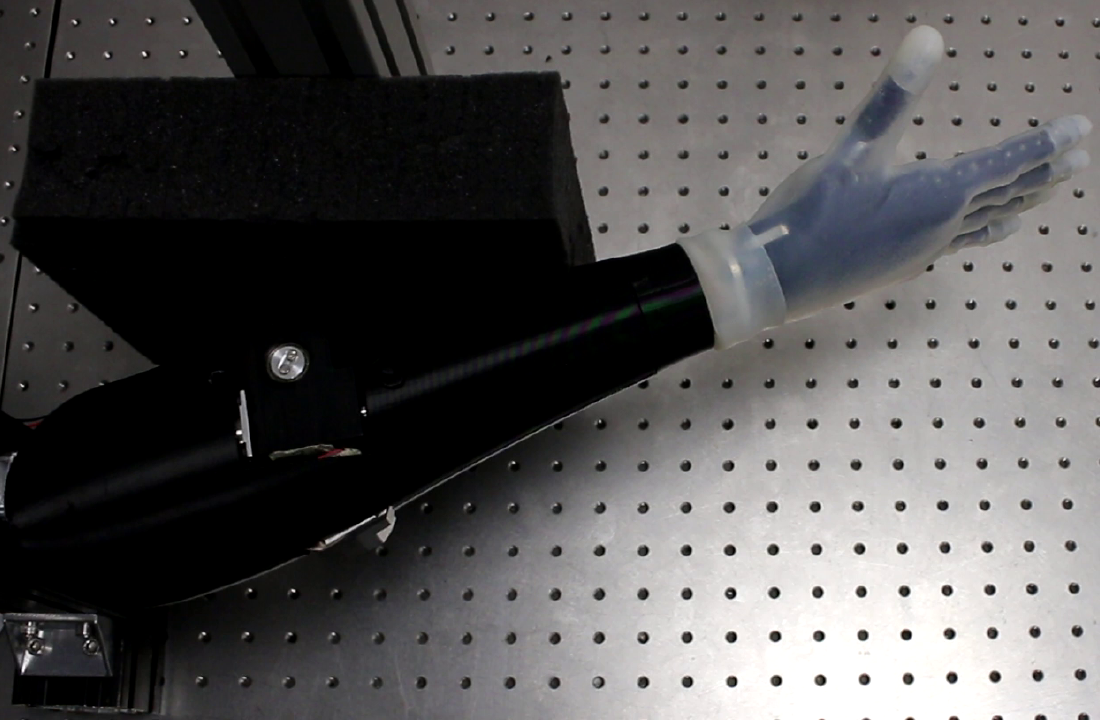}\label{fig:Obstacle_a}}
    \hfill
    \subfloat[]{\includegraphics[width = 0.49\linewidth]{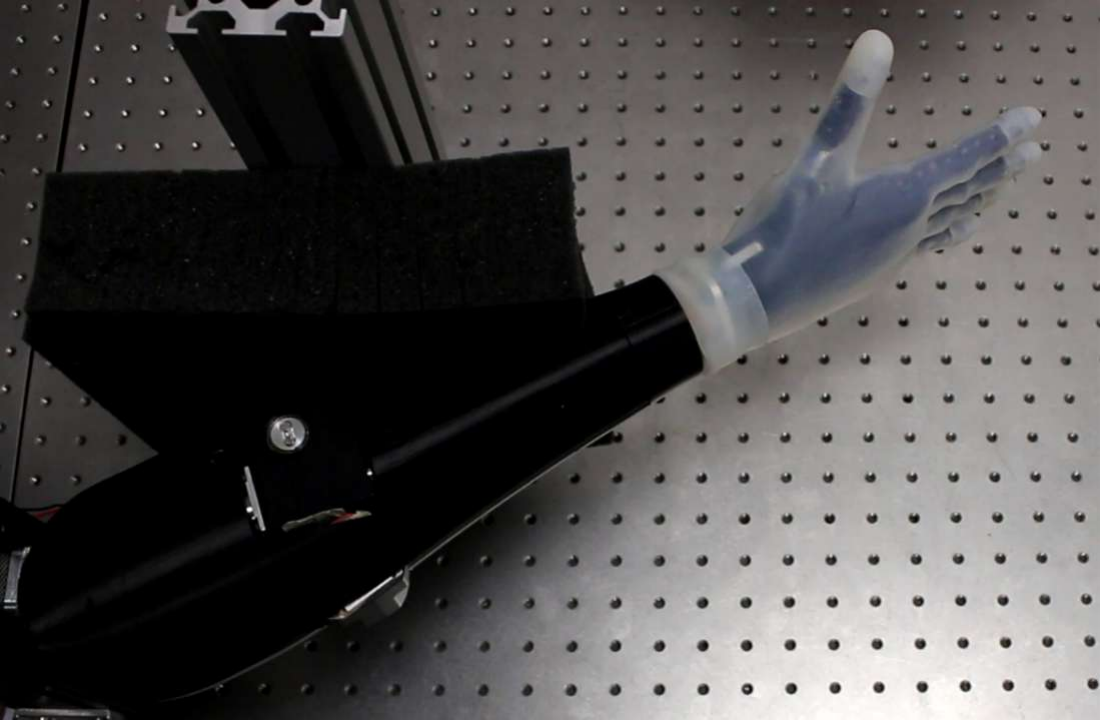}\label{fig:Obstacle_b}}\\
    \subfloat[]{\includegraphics[width = 0.98\linewidth]{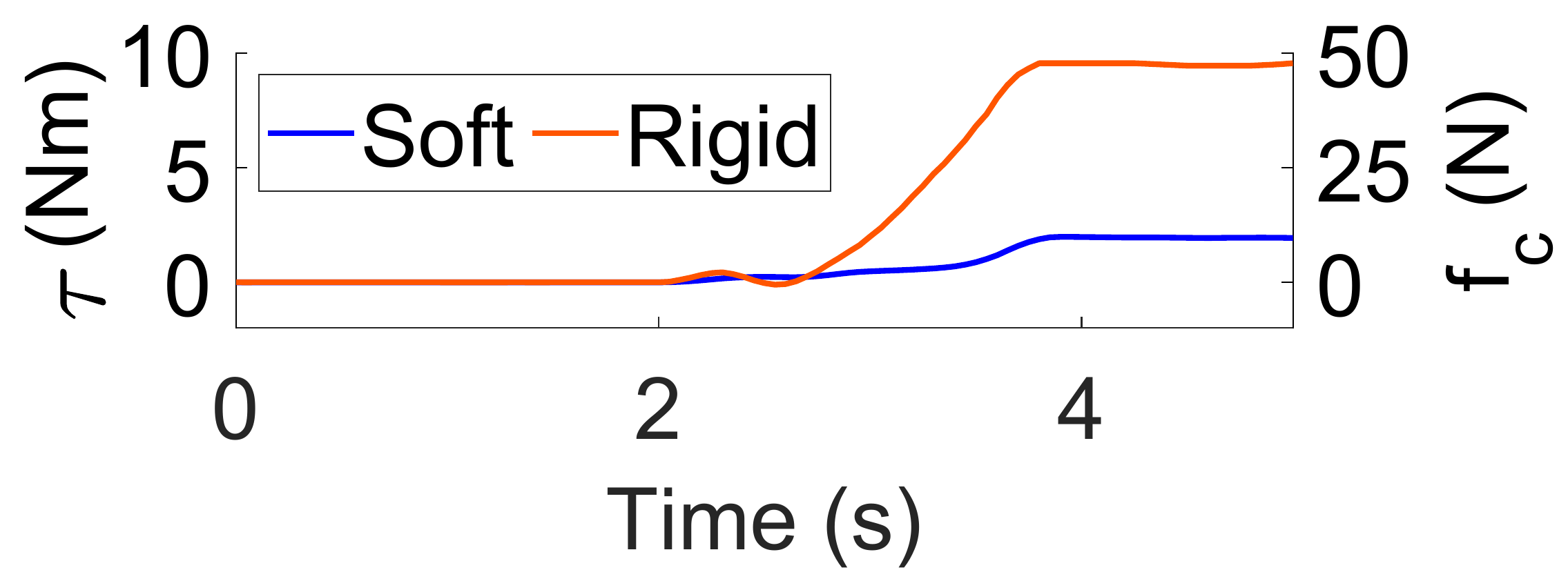}\label{fig:Obstacle_c}}
    \caption{The VS-Arm AA encountering a fixed obstacle during motion at low (a) and high (b) joint stiffness. Graph (c) illustrates the external torque $\tau$ and contact force $f_c$ applied by the elastic joint during the interaction with the fixed obstacle in the soft (blue line) and rigid (orange line) configurations. In the soft configuration (a), the elastic joint complies with the environment, resulting in a faint reaction torque. In contrast, rigid actuation (b) causes high interaction forces when environmental factors obstruct the motion.}
    \label{fig:Obstacle}
\end{figure}
\begin{figure}
    \centering
    \subfloat[]{\includegraphics[width = 0.49\linewidth]{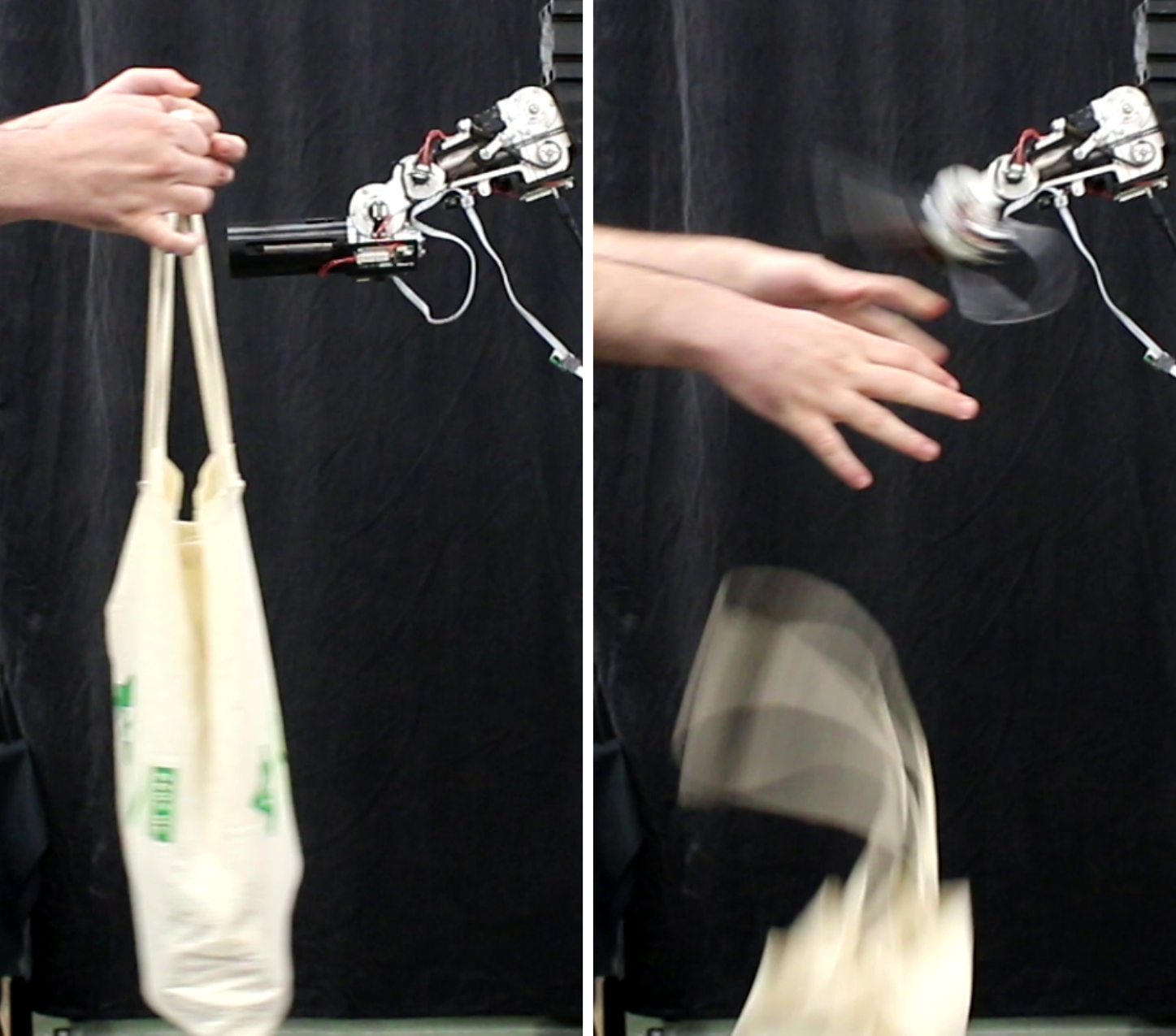}\label{fig:Bag_a}}
    \hfill
    \subfloat[]{\includegraphics[width = 0.49\linewidth]{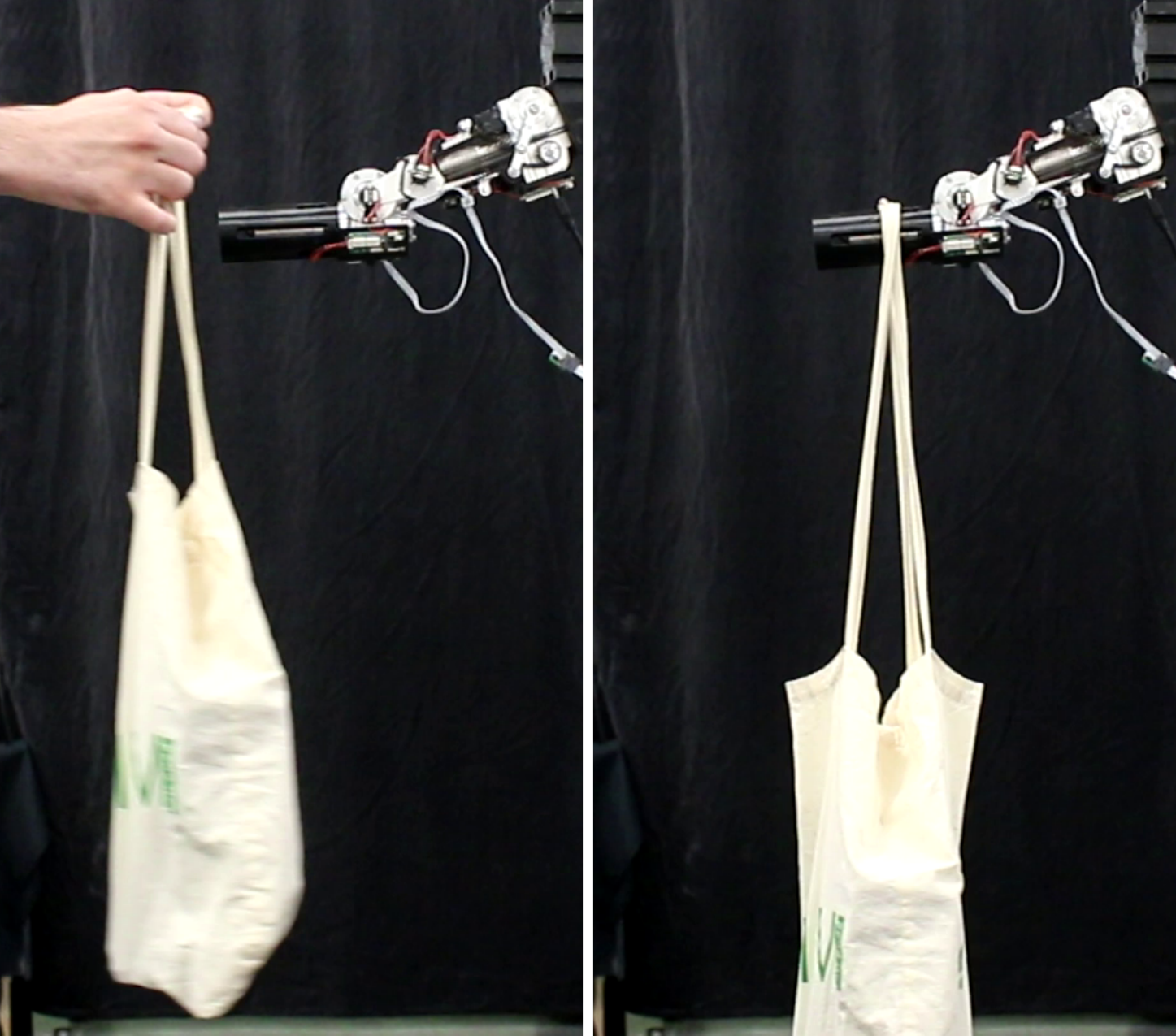}\label{fig:Bag_b}}
    \caption{The VS-Elbow D2 carrying a 2 kg bag. Adopting low elbow stiffness (a), the elastic joint deflects until dropping the bag. Conversely, the prosthesis can support the payload without significantly affecting its absolute position by exploiting high elbow stiffness (b).}
    \label{fig:Weight_Bag}
\end{figure}
The experimental validation highlighted some differences in the performance of the two architectures. These distinctions are mostly due to the diverse mechanical layouts (i.e., agonist-antagonist and independent) and the size of the actuation systems. 
We adopt an asymmetric motor layout in the VS-Elbow AA because the biceps motor unit is primarily responsible for counteracting gravity, while the triceps motor does not require as high torque. This asymmetrical motor choice reduces the overall size, weight, and power consumption of the prototype. However, in the AA layout, both motors are required to control the joint position and stiffness. Consequently, the less powerful motor (i.e., the triceps) limits the maximum output stiffness, while the slowest one (i.e., the biceps) bounds the joint velocity. Given that morphological factors are particularly relevant in prosthetic design, we believe the asymmetric motor choice provides a reasonable and advantageous trade-off between user comfort and performance.

Fig.~\ref{fig:Elastic_Exp} illustrates the experimental elastic behavior of the system. The elastic transmission achieves non-linear stiffness characteristics due to the geometry of the mechanism. Some VSAs decouple joint stiffness from deflection by employing quadratic springs \cite{migliore2005biologically} or by altering the transmission ratio between linear springs and the output shaft \cite{groothuis2013variable, kizilhan2015comparison}, thus maintaining constant stiffness even when the joint deflects. In contrast, with the proposed system, the joint stiffness varies with deflection even when the spring preload is fixed, providing a self-adaptive response to increasing perturbations. Based on current knowledge, it is unclear whether this behavior may benefit the user in final applications. However, human muscles notably feature non-linear torque and stiffness characteristics \cite{gribble1998complex, mghames2017design} that can be more faithfully emulated with the presented mechanism. The obtained stiffness curves result from the mechanical design of the elastic transmission. We defer to future investigation the potential parametric optimization to achieve a desired elastic characteristic, as done in \cite{baggetta2022design}.

Fig.~\ref{fig:Position_Step} and \ref{fig:Stiffness_Step} highlight the main differences between the agonist-antagonist and independent setups. 
In the distributed implementation, since the positioning and stiffness mechanisms are decoupled, the speed of the output link is independent of the stiffness configuration. In the AA setup, instead, the effects of different spring preloads are visible on the output joint position: low stiffness leads to faster response and oscillations, while high stiffness brings a slower but steadier reaction. The VS-Elbow D2 exhibited a larger static error in achieving the desired posture compared to the VS-Elbow AA, primarily due to differences in controller gain tuning. The D2 implementation uses a lower proportional gain to ensure smoother motion, whereas the AA device employs higher proportional gains to maintain accurate motion even with high internal preload.
Fig.~\ref{fig:Stiffness_Step} shows that the stiffness motor does not endure the external payload in the distributed implementation. 
In contrast, in the AA implementation, the external payload increases the SVT and introduces a static error in the stiffness control due to the resisting torque. During the experiment, the position reference is maintained to achieve maximum torque at maximum stiffness, ensuring full power for stiffness regulation. However, as the system transitions from a stiff to a soft configuration, increased deflection reduces the lever arm of the gravitational force, thereby decreasing the external torque on the elbow joint. This experiment highlighted the need for a posture compensation strategy, which was integrated into the system controller after the characterization experiments.

Table~\ref{tab:comp_elbow} compares the most relevant features of the presented system with the human and other prosthetic elbow joints.
Both versions of the prosthesis appear well-suited for prosthetic applications, as they meet the target requirements for an elbow prosthesis and offer comparable weight, size, and performance to existing devices. 
However, the proposed VS prosthesis necessitates an extra motor for the stiffness modulation, which results in a slight reduction in maximum torque to prioritize lightweight design. Despite this, the system can actively lift up to 3 kg, and its passive locking mechanisms support up to 5 kg without consuming power, unlike the backdrivable devices discussed in \cite{sensinger2008user, bennett2016design}. Although the weight of the presented system is significantly lower than other passive-VS elbows in the literature and comparable to commercial devices, further optimization may be necessary to reduce its weight and enhance user comfort in the final application.
Note that, apart from the prototypes presented here, in \cite{lemerle2019variable}, and in \cite{baggetta2022design}, all the other devices present fixed or software-controlled impedance. 

The case studies underscore the capability of VS elastic actuation to empower prostheses for safe interactions with the environment and adaptability to unpredictable scenarios.
The compliant nature of human muscles allows them to absorb sudden impacts effectively. In contrast, conventional prostheses with rigid actuation systems lack this compliance, thus being potentially harmful in unstructured environments like the real world \cite{haddadin2007safety}. Fig.~\ref{fig:Hammer} illustrates the system response to collisions at different stiffness configurations. Notably, the elastic joint deflects and safely absorbs shocks in the soft configuration. Conversely, high joint stiffness enables the system to resist perturbations but results in substantial interaction forces, posing risks to people and objects. While the main message of the experimental results can be observed using manually produced impacts, the accuracy of the numerical outcomes is limited due to the uncertain repeatability of the impacts.
\begin{figure}
    \centering
    \includegraphics[width = 0.98\linewidth]{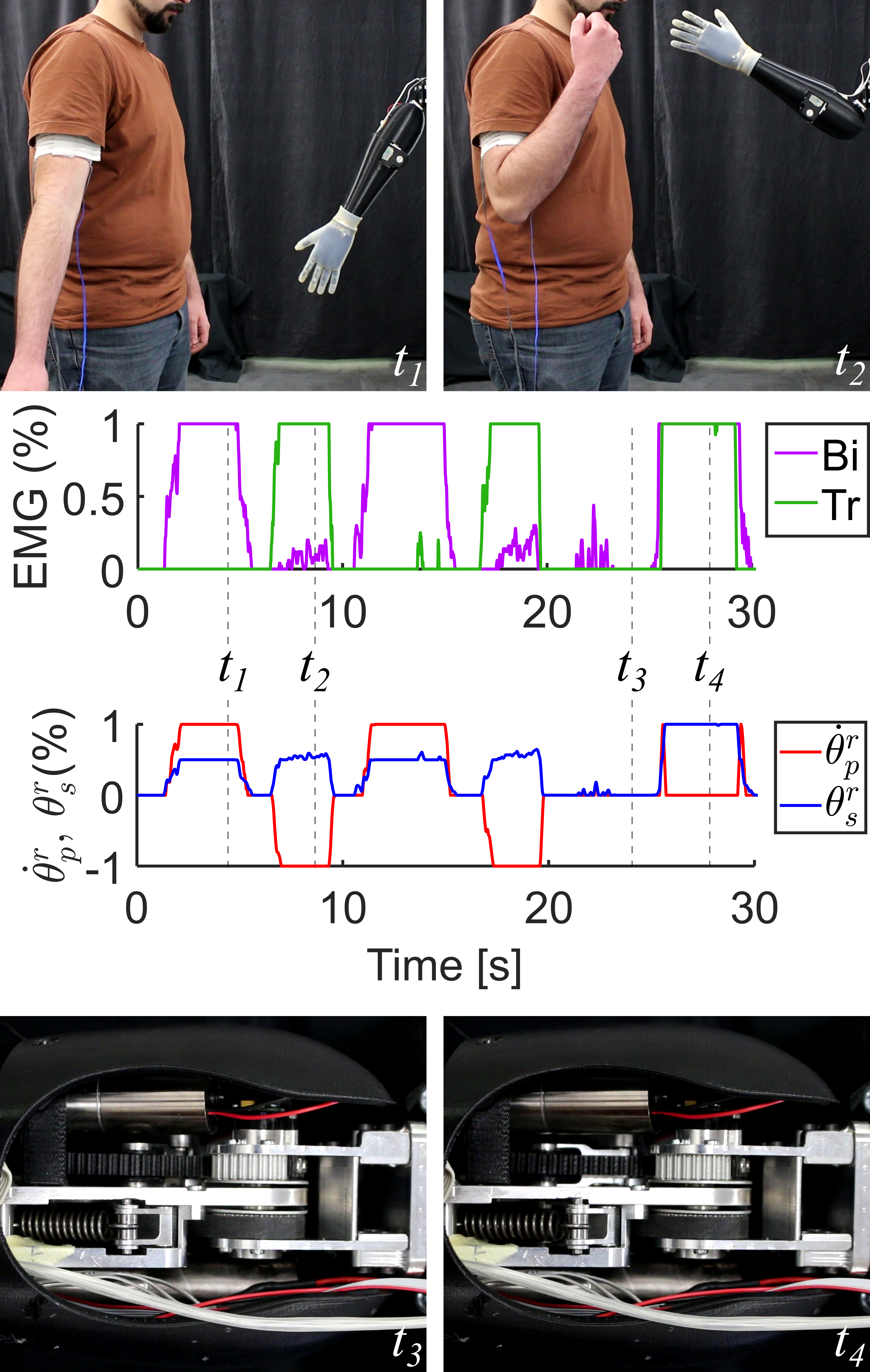}
    \caption{Healty subject controlling the VS-Elbow AA using a pair of surface EMG electrodes. The first graph illustrates the normalized EMG activity of the biceps (in magenta) and triceps (in green) brachii. The second one depicts the normalized joint speed (in red) and stiffness (in blue) control references obtained by implementing the VS EMG controller described in \cite{capsi2020exploring}. The photos portray distinct time frames during the experiments, indicated by labels between the graphs. The prosthesis correctly mimics the user's elbow flexion ($t_1$) and extension ($t_2$), continuously adapting its impedance proportionally to the average level of muscular contraction. Upon detecting a pure coactivation, the prosthesis halts its motion, transitioning from a compliant ($t_3$) to a rigid ($t_4$) configuration.}
    \label{fig:EMG_control}
\end{figure}
Rigid actuation systems can also be perilous during active interactions in unstructured environments. Prostheses often lack haptic feedback, thus the users must rely solely on visual cues. If the user does not promptly detect contacts during movements, rigid actuation can rapidly generate high interaction forces before they can halt the motion command.  
Fig.~\ref{fig:Obstacle_c} illustrates that without prompt intervention by the user, a rigid system generates very high torque in less than a second. In contrast, low joint stiffness enables users to safely interact with the environment by exerting small contact forces. 
Similarly, the supplemental multimedia material demonstrates that low joint stiffness prevents the prosthesis from accidentally knocking down objects during motion when they are not fixed.
Nevertheless, Fig.~\ref{fig:Weight_Bag} reveals that high joint stiffness is crucial for supporting a moderate payload without causing significant deviations in joint position. Therefore, a VS prosthesis becomes imperative to tailor bionic limb impedance to accommodate daily tasks.
Fig.~\ref{fig:EMG_control} shows that the EMG control strategy effectively conveys the user's stiffness and motion intentions within the bionic limb. However, slight motions may arise during pure cocontractions if the muscular activity of antagonist muscles is not perfectly synchronized. Moreover, the algorithm limits fine stiffness modulation during active motion. Therefore, further efforts are needed in developing VS EMG controllers.

%% file: Sections/Conclusions.tex
\section{Conclusions} \label{sec:conclusions}
This work focuses on developing a novel elbow prosthesis with variable joint stiffness to enhance the robustness, safety, and adaptability of the device in unstructured environments. The user-controllable impedance is achieved passively by reconfiguring the spring preload through redundant elastic actuation.
The presented system consists of two distinct mechanical architectures to cater to the specific needs of different users. Drawing inspiration from nature, the VS-Elbow AA can be entirely contained in the user's upper limb to mimic the missing biceps and triceps brachii by utilizing an AA actuation layout. While this implementation saves space in the forearm for additional prosthetic components, such as a multi-DoF wrist or tactile feedback devices, it concentrates most of the device weight in the upper limb, potentially causing discomfort. To address this concern, a novel VSA with sparse motor allocation was developed and adopted in the VS-Elbow D2, achieving homogeneous and human-like mass distribution. However, such architectures may not be suitable for distal transhumeral amputations since they allocate components in the upper limb segment. Addressing this limitation, we present an alternative solution that adopts the AA implementation to create a complete transhumeral prosthesis, locating all its components below the elbow joint and thereby being suitable for fitting lengthy residual limbs. While the independent motor layout is more intuitive for the distributed implementation and the AA architecture for the concentrated device due to the intrinsic structure of their actuation systems, it is also feasible to reverse these technological solutions to accommodate different levels of amputations.
The experimental characterization and validation confirmed that the presented system meets the target requirements for prosthetic applications. The reported case studies offer compelling evidence of the significant benefits that VSAs provide to prosthesis users, including more natural interactions that enhance users' sense of belonging and quality of life.

Future efforts will be directed toward improving the user interface and comprehensively assessing the benefits of stiffness modulation in functional tasks with prosthesis users. While existing literature already includes simple EMG controllers for single-DoF VS prostheses \cite{sensinger2008user, capsi2020exploring}, more advanced control strategies may be necessary to enable users to exploit the controllable impedance of their artificial limbs fully.